\documentclass[10pt,twocolumn,letterpaper]{article}

\usepackage{cvpr}              

\definecolor{cvprblue}{rgb}{0.21,0.49,0.74}
\usepackage[pagebackref,breaklinks,colorlinks,allcolors=cvprblue]{hyperref}

\def\paperID{7029} 
\def\confName{CVPR}
\def\confYear{2025}
\usepackage{hyperref}
\usepackage{array}
\usepackage{multirow}
\usepackage{colortbl} 
\usepackage{xcolor} 
\title{From Zero to Detail: Deconstructing Ultra-High-Definition Image Restoration from Progressive Spectral Perspective} 

\author{
  Chen Zhao$^{1*}$, \hspace{0.2cm}
  Zhizhou Chen$^{1*}$, \hspace{0.2cm}
  Yunzhe Xu$^1$, \hspace{0.2cm}
  Enxuan Gu$^2$, \hspace{0.2cm}
  Jian Li$^3$ \\
  Zili Yi$^1$, \hspace{0.2cm}
  Qian Wang$^4$, \hspace{0.2cm}
  Jian Yang$^{1}$, \hspace{0.2cm}
  Ying Tai$^{1\dagger}$ \\
  $^1$Nanjing University, \hspace{0.2cm}
  $^2$Dalian University of Technology, \hspace{0.2cm}
  $^3$Tencent Youtu, \hspace{0.2cm}
  $^4$China Mobile Institute 
}

\begin{document}
\maketitle
\vspace{-1cm}
\begin{abstract}
	Ultra-high-definition (UHD) image restoration faces significant challenges due to its high resolution, complex content, and intricate details. To cope with these challenges, we analyze the restoration process in depth through a progressive spectral perspective, and deconstruct the complex UHD restoration problem into three progressive stages: zero-frequency enhancement, low-frequency restoration, and high-frequency refinement. Building on this insight, we propose a novel framework, ERR, which comprises three collaborative  sub-networks: the zero-frequency enhancer (ZFE), the low-frequency restorer (LFR), and the high-frequency refiner (HFR). Specifically, the ZFE integrates global priors to learn global mapping, while the LFR restores low-frequency information, emphasizing reconstruction of coarse-grained content. Finally, the HFR employs our designed  frequency-windowed kolmogorov-arnold networks (FW-KAN) to refine textures and details, producing high-quality image restoration. Our approach significantly outperforms previous UHD methods across various tasks, with extensive ablation studies validating the effectiveness of each component. The code is available at \href{https://github.com/NJU-PCALab/ERR}{here}.

    \renewcommand{\thefootnote}{}
    \footnotetext{$^*$Equal contributions.  $\dagger$ indicates corresponding author.}
    \renewcommand{\thefootnote}{\arabic{footnote}}

\end{abstract}
\vspace{-0.5cm}
\section{Introduction}
\vspace{-0.2cm}
Ultra-high-definition (UHD) imaging has advanced rapidly, and has gained widespread attention across diverse areas \cite{li2023embedding,yu2022towards,zheng2021ultra,deng2021multi,li2023uhdnerf,sun2024ultra,sunprogram}. However, UHD images captured under adverse conditions, such as low light, rain, or haze, often suffer from severe degradation \cite{wang2024correlation}. The purpose of this paper is to explore UHD image restoration (IR).

With the remarkable success of deep learning \cite{DBLP:journals/corr/abs-2408-05205,DBLP:conf/eccv/JinLWZZ24,DBLP:conf/cvpr/LiLZJF0024,DBLP:journals/nn/ZhaoCHY24,DBLP:journals/corr/abs-2403-01497,DBLP:journals/corr/abs-2404-01717,DBLP:journals/corr/abs-2411-06558,DBLP:conf/cvpr/ZhouLMZY24,DBLP:journals/pami/ZhouLMYY25,DBLP:conf/iccv/LuLK23,DBLP:conf/cvpr/LuWLLK24,DBLP:journals/pami/TianLYZCG24,DBLP:journals/ijcv/TianYZGG22,DBLP:journals/tip/TianYZCG23,DBLP:conf/iccv/0017LZG23,DBLP:journals/corr/abs-2408-17135,zhangrethinking,zhang2024vocapter,zhang2025u,zhang2024catmullrom}, a wide range of learning-based methods have been proposed to address IR challenges, focusing on convolutional neural networks (CNNs) and Transformer \cite{tai2017memnet,zhang2017learning,liang2021swinir,zamir2022restormer,chen2023learning,zhou2024adapt,zhang2019residual,DBLP:journals/ijcv/WangZSLSLKLL24,DBLP:conf/eccv/SunRGWC24,cheng2024spt}. 
Current state-of-the-art (SOTA) methods primarily enhance network performance by introducing more complex architectures \cite{li2023efficient,zhangxformer,chen2024bidirectional,zhang2024distilling,DBLP:conf/cvpr/0002FZL00Z24,ye2024learning,qin2024restore,zhou2023pyramid,DBLP:conf/iccv/ShaoXWG0H23}. However, due to the ultra-high resolution and pixel density of UHD images, these advanced methods struggle to perform effectively in UHD scenarios.


\begin{figure}[t]
	\centering
	\includegraphics[width=0.98\linewidth]{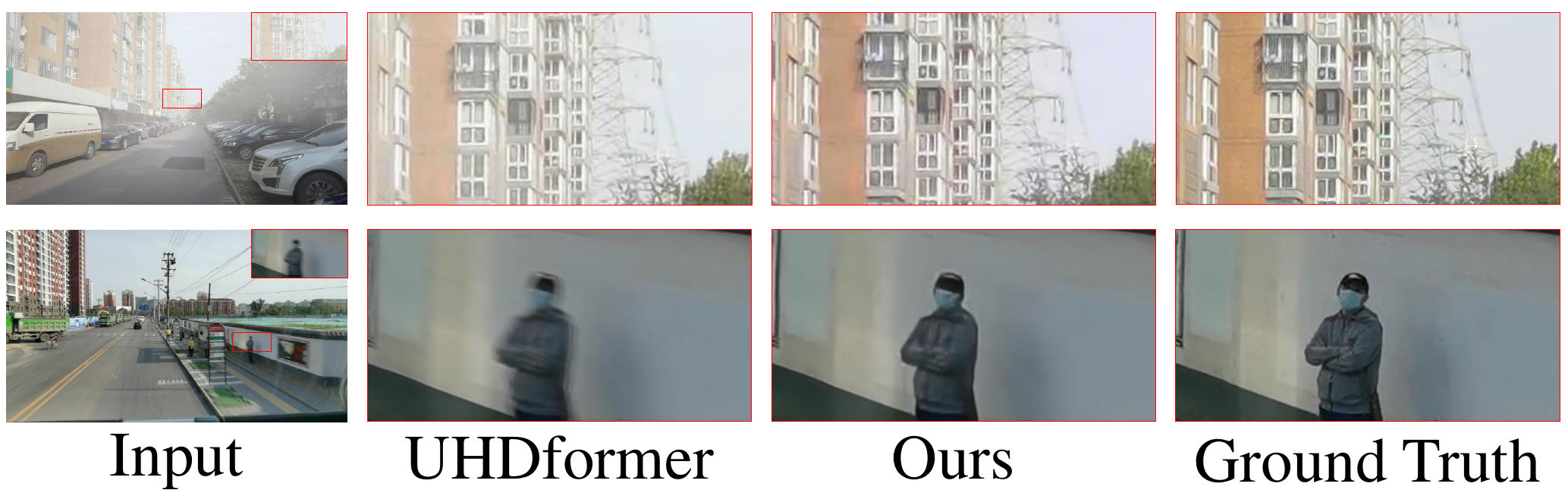}
    \vspace{-0.2 cm}
	\caption{Visual comparison with the latest SOTA method.  } 
	\vspace{-0.7 cm}
	\label{fig:0}
\end{figure}

Recently, several methods tailored for UHD IR have emerged \cite{wang2024correlation,zou2024wave,wang2023ultra,DBLP:conf/mm/YuZZHZZ23,DBLP:conf/mm/XiaoLW24,wang2024public}. LLformer \cite{wang2023ultra} achieved impressive performance by leveraging Transformers. However, due to the high computational cost, this approach cannot efficiently perform full-resolution inference on edge devices. UHDFour \cite{li2023embedding} reduced the resolution  through 8× downsampling, enabling full-resolution inference of UHD images on edge devices.  UHDformer \cite{wang2024correlation} proposed a correction transformer that leverages high-resolution feature to guide the low-resolution restoration. Despite these methods relying on downsampling reduced computational costs, this \textit{downsampling-enhancement-upsampling learning paradigm} inevitably leads to the loss of critical information \cite{yu2024empowering}. Furthermore, the complexity of UHD images, characterized by \textit{ ultra-high resolution, substantial content, and intricate structural details,} poses significant challenges for restoration, making it  difficult for existing methods to achieve efficient and high-quality results \cite{yu2024empowering}.


\begin{figure*}[t]
	\centering
	\begin{subfigure}[t]{0.82\linewidth}
		\centering
		\includegraphics[width=\linewidth]{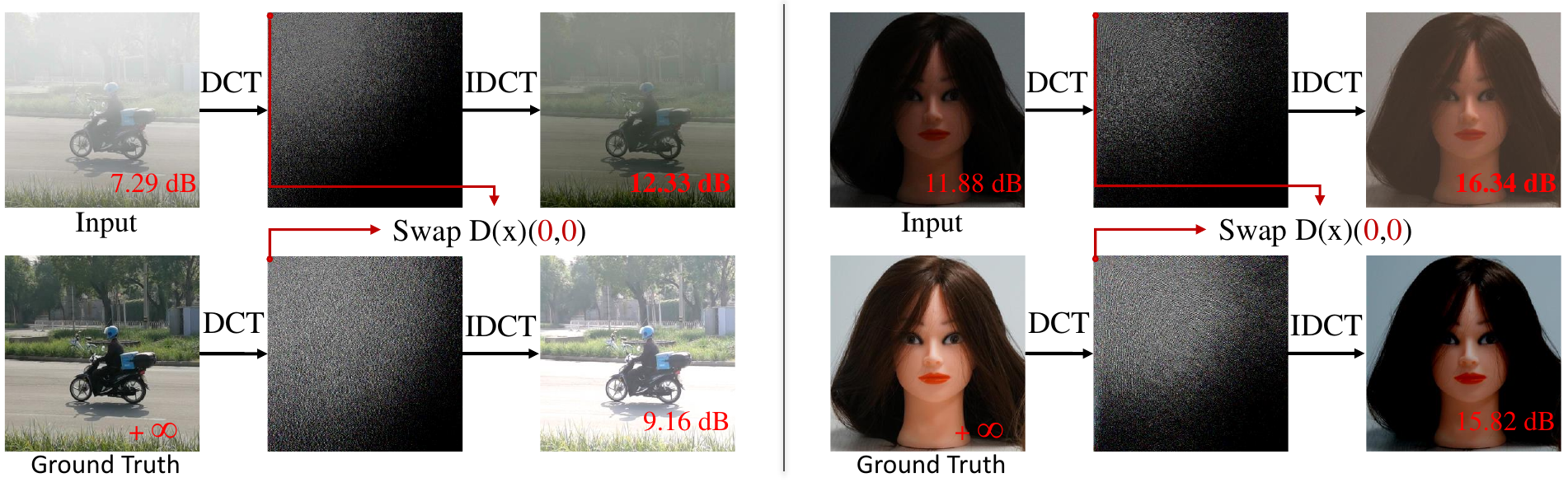}
		\caption{
        We exchange the information at the (0,0) position in the DCT spectrum, which represents the global information.
		}
	\label{fig:2a}
	\end{subfigure}%
	
	\vskip 0.1cm  

	\begin{subfigure}[t]{0.95\linewidth}
		\centering
		\includegraphics[width=\linewidth]{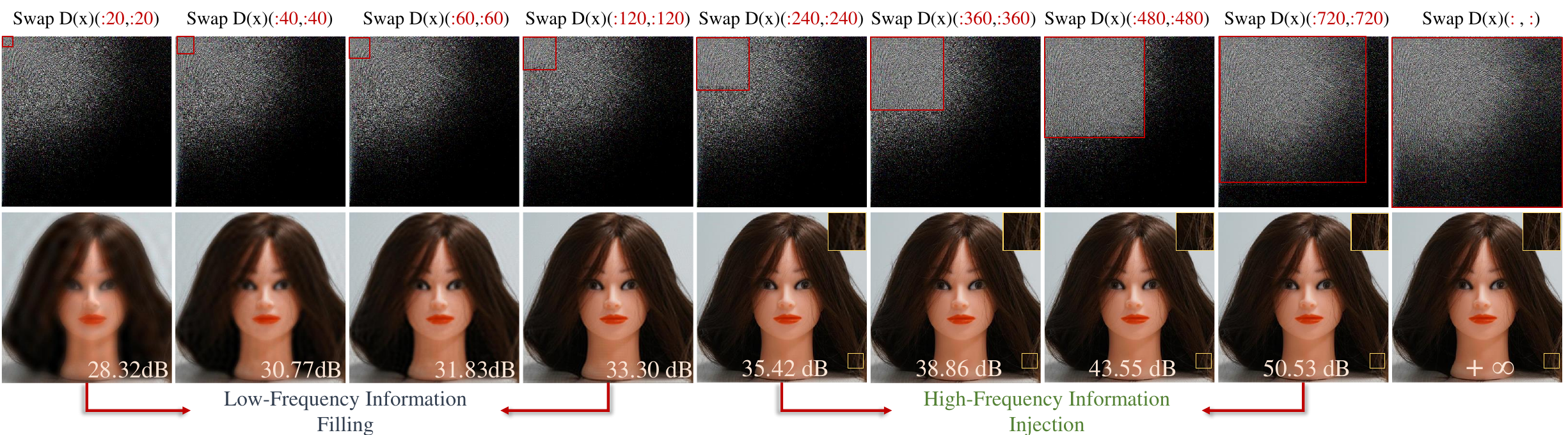}
		\caption{
       The low-frequency filling restores the coarse-grained 
content, while the high-frequency injection refines the fine-grained textures.
        }
	\label{fig:2b}
	\end{subfigure}
	
	\vspace{-0.3cm}
	\caption{Our core motivation. Based on the observations in (a) and (b), we deconstruct the complex UHD restoration problem into three progressive stages: zero-frequency enhancement, low-frequency restoration, and high-frequency refinement.
    }
    \vspace{-0.5cm}
	\label{fig:2}
\end{figure*}

To investigate the issue of UHD IR in depth, we adopt a progressive frequency decoupling to examine the significance of various frequency components in restoration process. Initially, we map the degraded image (input) and the corresponding ground truth (GT) into the frequency domain using discrete cosine transform (DCT), exchanging the frequency information at the (0,0) position, known as the \textit{zero frequency component} \footnote{\href{https://web.stanford.edu/class/archive/engr/engr40m.1178/slides/signals.pdf}{stanford.edu/signals.pdf}}.
We then reconstruct the exchanged frequency spectrum back into the spatial domain using inverse discrete cosine transform (IDCT), as illustrated in Figure \ref{fig:2a}. The zero-frequency component represents the direct-current information, reflecting the global and average characteristics of the image. We observe that the visual properties  of the exchanged input and GT also swaps, and the PSNR value of the exchanged input slightly higher than that of the exchanged GT. This suggests that even if perfect non-zero frequency component is learned (e.g., the exchanged GT), acceptable results cannot be achieved, further indicating that \textit{the zero-frequency component plays a crucial role in the early stages of restoration.} we progressively expand the range of exchanged frequency components and observe that as low-frequency component is filled, the structures and content are restored, while the injection of high-frequency information refines the details and textures, as shown in Figure \ref{fig:2b}. 
\textit{Based on the observation, we deconstruct the complex UHD IR problem into three progressive stages: zero-frequency enhancement, low-frequency restoration, and high-frequency refinement, each targeting to learn the global mapping, coarse-grained
content, and fine-grained textures, respectively, thereby enhancing the  quality of images}, as depicted in Figure \ref{fig:0}.

Building on the insight, we develop a novel framework, termed ERR, which leverages progressive frequency decoupling to navigate the complexities of UHD images.
ERR consists of three collaborative sub-networks: the zero-frequency enhancer (ZF\textbf{E}), the low-frequency restorer (LF\textbf{R}), and the high-frequency refiner (HF\textbf{R}). Specifically, by incorporating global prior, the ZFE focuses on enhancing global information. The purpose of the LFR is to further restore the low-frequency information, emphasizing the reconstruction of the primary content. Finally, the HFR employs our meticulously designed frequency windowed kolmogorov-arnold networks (FW-KAN) to refine details and textures, achieving high-quality image restoration. Comparative results across multiple tasks indicate that our developed ERR achieves significant superiority over previous UHD methods. Furthermore, extensive ablation studies prove the effectiveness of our contributions.

\begin{figure*}[t]
	\centering
	\includegraphics[width=0.95\linewidth]{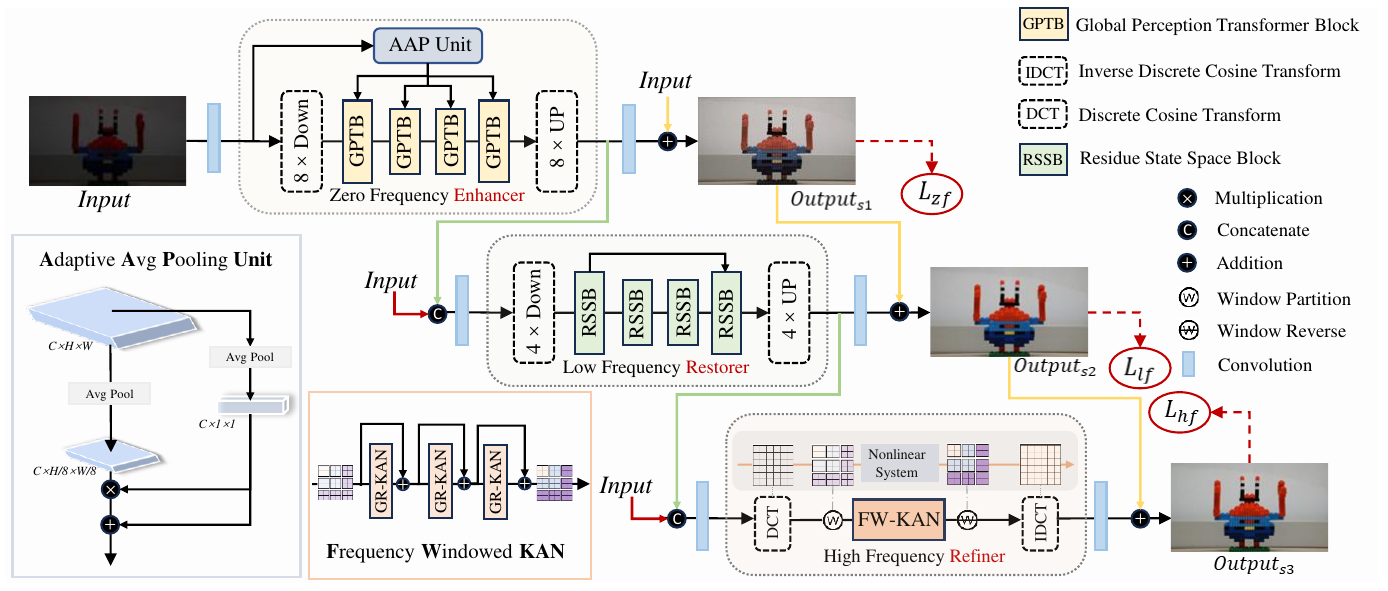}
	\caption{Overall framework of our proposed ERR. Building on the insight from progressive spectral perspective, ERR consists of three collaborative sub-networks: the zero-frequency 
enhancer (ZFE), the low-frequency restorer (LFR), and the high-frequency refiner (HFR).  } 
	\vspace{-0.4cm}
	\label{fig:3}
\end{figure*}

In summary, the key contributions of our work are as follows:

\begin{itemize}
	\item We conduct an in-depth analysis of UHD restoration from a progressive spectral perspective, deconstructing the complex UHD restoration problem into three progressive stages: zero-frequency enhancement, low-frequency restoration, and high-frequency refinement.
	
	\item Building on this insight, we propose a novel framework, termed \textbf{ERR}, which consists of three sub-networks: the zero-frequency enhancer (ZF\textbf{E}), the low-frequency restorer (LF\textbf{R}), and the high-frequency refiner (HF\textbf{R}).
	
	\item For the ZFE, we design a global perception transformer block (GPTB) to more effectively capture global representations. For the HFR, we develop a frequency-windowed KAN (FW-KAN) to refine fine-grained information, thereby enhancing image details and textures.

\end{itemize}
\vspace{-1cm}

\section{Related Works} \label{sec:related_work}

\subsection{Ultra-high-definition Image Restoration }

As UHD imaging  becomes more pervasive, the field of UHD restoration is drawing heightened attention \cite{wang2023ultra,zhuo2021ultra,yu2022towards,zheng2021ultra,deng2021multi,li2023uhdnerf,wang2024correlation,zou2024wave,wei2023efficient}. 
Zheng et al. \cite{zheng2021ultra} introduced a multi-guided bilateral upsampling model for UHD dehazing. UHDFour \cite{li2023embedding} downsampled UHD images by a factor of 8, enabling full-resolution inference on edge devices.  UHDformer \cite{wang2024correlation} leverages high-resolution information to guide the low-resolution restoration. UDR-Mixer \cite{chen2024towards} facilitated low-resolution spatial feature restoration through frequency feature modulation. These methods \cite{zheng2021ultra,li2023embedding,wang2024correlation,chen2024towards} share a similar paradigm: learning from downscaled images to ease computational demands. However, this \textit{downsampling-enhancement-upsampling paradigm} inevitably results in the loss of essential information \cite{yu2024empowering}. Moreover, the complexity of UHD images—with \textit{ultra-high resolution, rich content, and complex structures}—poses major challenges for restoration, making it difficult for existing methods to achieve both efficiency and high quality \cite{yu2024empowering}.

\subsection{Frequency Learning}
An increasing number of studies \cite{qin2021fcanet,fritsche2019frequency,mao2023intriguing,zou2024wave,lv2024fourier,mao2024loformer,zhou2024general,zhao2024toward,cui2023image,DBLP:journals/corr/abs-2410-12669,DBLP:journals/corr/abs-2410-18775,DBLP:conf/icmcs/LouZXCW024,DBLP:conf/bibm/LouXZZCW024,zhao2024spectral} leverage frequency domain characteristics across various tasks. 
DSGAN \cite{fritsche2019frequency}
separated the low and high image frequencies via the low- and high-pass filters. 
LaMa \cite{suvorov2022resolution} uses the frequency
convolution for image inpainting. DeepRFT \cite{mao2023intriguing} proposed a simple res-fft-relu-block for image deblurring.  Fourmer \cite{zhou2023fourmer} leverages the Fourier transform to model global dependencies for image restoration. To exploit the characteristics of different frequency signals, recent works \cite{zhao2024wavelet,jiang2023low,jiang2023dawn,zou2024wave} have explored discrete wavelet transform (DWT) to achieve decoupling purposes for IR tasks. Unlike previous frequency-domain approaches, we deconstruct the complex UHD restoration problem into three stages from a progressive frequency perspective,  focusing on learning the global mapping, coarse-grained 
content, and fine-grained textures, respectively. 


\vspace{-0.4cm}
\section{Methodology}
\vspace{-0.3cm}
\subsection{Overall Framework}
Given a UHD degraded image as input, we aim to learn a network to generate a UHD output that eliminates the dagradation. The framework of ERR is shown in Figure \ref{fig:3}, which integrates three collaborative sub-networks: the zero-frequency enhancer (ZFE), the low-frequency restorer (LFR), and the high-frequency refiner (HFR). Specifically, ZFE learns global representations within low-resolution space. To efficiently learn content representations, the LFR restores low-frequency information of the degraded UHD image within mid-resolution space. In the final stage, the HFR employs our designed frequency windowed kolmogorov-arnold networks (FW-KAN) to refine fine-grained information within full-resolution space, enhancing image details and textures to further achieve high-quality image reconstruction.
\begin{figure}[t]
	\centering
	\includegraphics[width=0.98\linewidth]{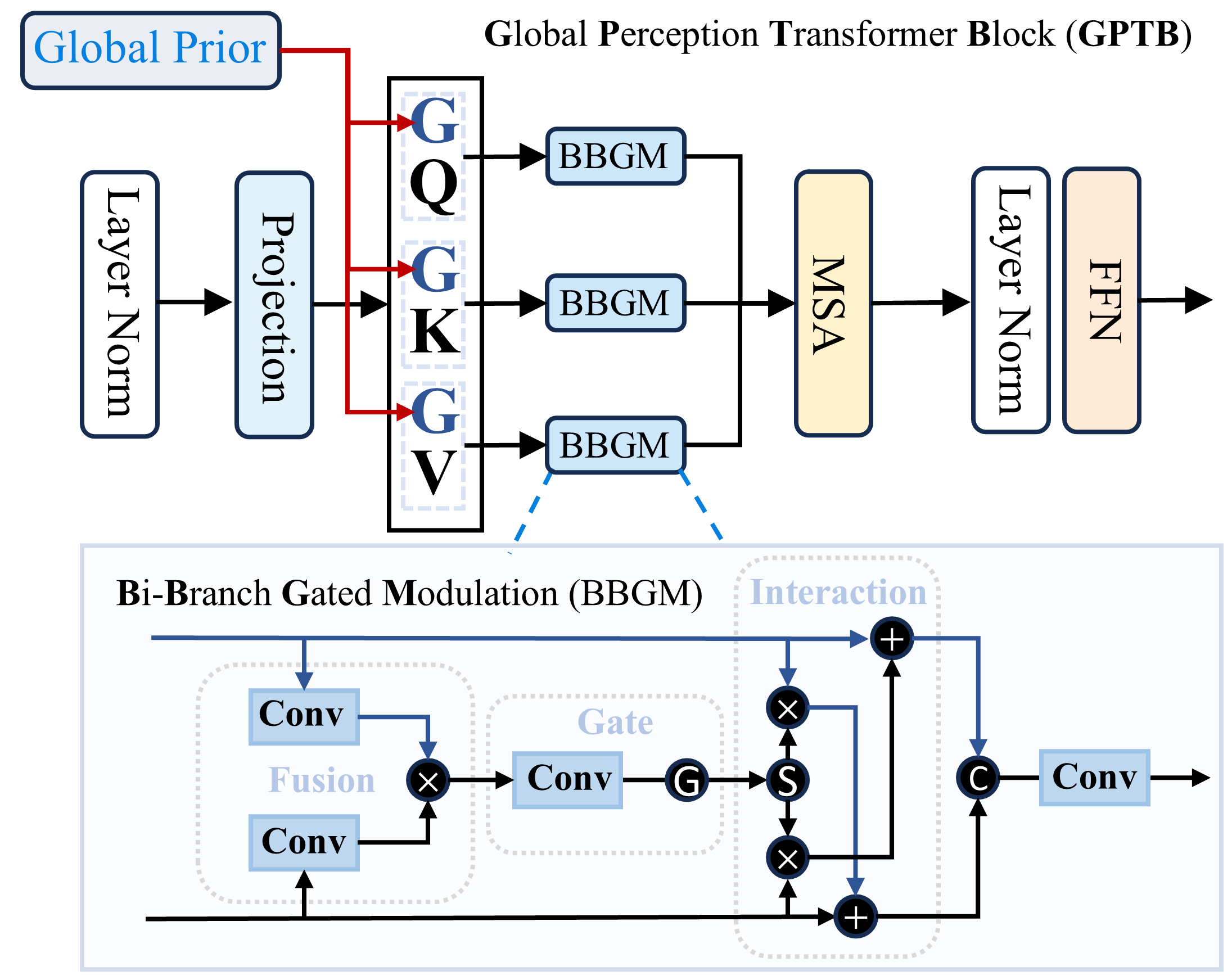}
     \vspace{-0.8em}
	\caption{The architecture of the Global Perception Transformer Block (GPTB).  } 
	\vspace{-0.5cm}
	\label{fig:4}
\end{figure}

\vspace{-0.3cm}
\subsection{Discrete Cosine Transform} 
Frequency decoupling and analysis have been widely applied across various fields. We primarily use the discrete cosine transform (DCT) to analyze and deconstruct UHD images. Given a input $x \in \mathbb{R}^{H\times W\times C}$, whose spatial shape is $H\times W$, the DCT transform $\mathcal{D}$ which converts the input $I$  to the
frequency space $F$ can be expressed as:
{
\footnotesize
\begin{align}
   \mathcal{D}(x)(u,v) &= F(u,v) \notag \\
   &= \sum_{h=0}^{H-1} \sum_{w=0}^{W-1} x(h,w) \cdot \cos\frac{(2h+1)u\pi}{2W} \cdot \cos\frac{(2w+1)v\pi}{2H},
\end{align}
}where $h, w$ and $u, v$ stand for the coordinates in the RGB color space and in the frequency space. $\mathcal{D}^{-1}$ denotes the inverse discrete cosine transform (IDCT).

\noindent \textbf{The core motivation}. To cope with the complexities of UHD images, we analyze and explore from progressive frequency decoupling perspective. Figure \ref{fig:2} illustrates our motivations schematically. Firstly, we investigate the importance of the zero-frequency component, which can be represented as $\mathcal{D}(x)(0,0)$. The zero-frequency component refers to the direct-current information, reflecting the global and average characteristics of the image. We exchange the zero-frequency components between the degraded image (input) and the GT, obtaining the exchanged input and GT. The exchanged input retains a perfect zero-frequency component while containing degraded non-zero frequency elements, while the exchanged GT is its inverse. As shown in the two cases in Figure \ref{fig:2a}, the quality of the exchanged input is significantly higher than that of the exchanged GT. Therefore, even if perfect non-zero frequency components can be learned (e.g., the exchanged GT), satisfactory results cannot be achieved. This observation suggests that \textit{in the early stages of restoration, a stronger focus should be placed on learning the zero-frequency component, i.e., capturing the global mapping}. We sequentially expand the range of frequency components, exchanging the first k levels, which can be expressed as $\mathcal{D}(x)(:k,:k)$. It is evident that we can observe two trends in the changes of image quality. As the low-frequency information is progressively filled, the content and structural information of the coarse-grained level are gradually restored, while the injection of high-frequency information refines the details and textures of the fine-grained level. The PSNR value and visual quality of the input gradually improve, as illustrated in Figure \ref{fig:2b}. This step-by-step  process from zero to detail 
implies a gradual enhancement in image quality; in contrast, the reverse process fails to achieve high-quality results. 



Based on this observation and analysis, our goal is to deconstruct the complex UHD restoration problem into three progressive stages: \textit{zero-frequency enhancement, low-frequency restoration, and high-frequency refinement, each focusing on learning the global mapping, coarse-grained content, and fine-grained textures, respectively.} 
\vspace{-0.8em}
\subsection{Zero Frequency Enhancer }
 ZFE aims to learn global mapping within low-resolution space, primarily through two core modules: the adaptive average pooling (AAP) unit and the global perception transformer block (GPTB). The purpose of the AAP unit is to extract global prior  within the full-resolution space. By leveraging the captured global prior, the GPTB can more effectively enhance global representation learning. 

\noindent \textbf{AAP unit}. The zero-frequency component represents the direct current (DC) information, which reflects the global and average characteristics of the UHD image. Consequently, we employ average pooling (AvP) in the full-resolution space to capture global prior. Given an input feature map $x\in\mathbb{R}^{H\times W\times C}$, we aim to extract global features while preserving local characteristics. To achieve the target, we employ a combination of global AvP and local AvP. The AAP unit can be expressed mathematically as:
{\begin{align}
   G=\mathcal{AAP}(x) = AvP_{1,1}(x) \odot AvP_{\frac{H}{8}, \frac{W}{8}}(x) + AvP_{1,1}(x),
\end{align}
}where $AvP_{1,1}$ represents the global AvP with a pooling size of \( (1, 1) \), and 
$AvP_{\frac{H}{8}, \frac{W}{8}}(x)$ denotes the local AvP with a pooling size of \( \left( \frac{H}{8}, \frac{W}{8} \right) \). Moreover, to enhance feature representation capacity, we introduce multi-scale learning within our AAP unit. $\odot$  indicates the element-wise  multiplication. $G \in\mathbb{R}^{\frac{H}{8}\times\frac{W}{8}\times C}$ refers to the global prior.

\begin{figure}[t]
	\centering
	\includegraphics[width=0.98\linewidth]{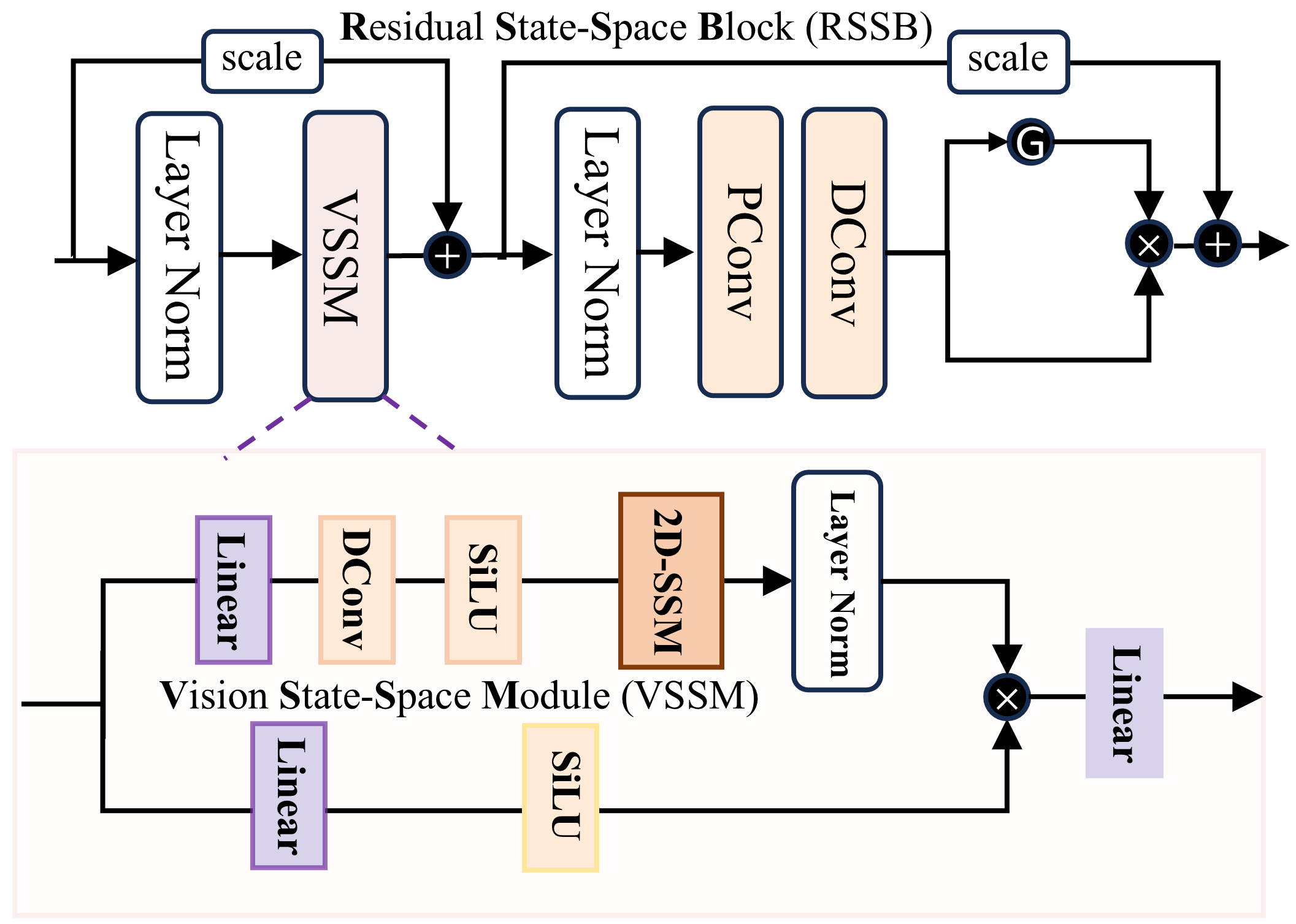}
     \vspace{-0.8em}
	\caption{The architecture of the residue state space block (RSSB). } 
	\vspace{-0.5cm}
	\label{fig:5}
\end{figure}
\noindent \textbf{GPTB}. The low-resolution features $x_{\mathrm{low}}\in\mathbb{R}^{\frac{H}{8}\times\frac{W}{8}\times C}$  are obtained from an input $x$ via 8x downsampling. Then, $x_{\mathrm{low}}\in\mathbb{R}^{\frac{H}{8}\times\frac{W}{8}\times C}$ is sent to several GPTB to learn the global mampping. Figure \ref{fig:4} shows the detail architecture of GPTB, where bi-branch gated modulation (BBGM) is designed to integrate low-resolution features with global priors.  The computation can be denoted in the GPTB as: 
\begin{align}
   Q,K,V= \mathcal{S}(Projection(LN(x_{\mathrm{low}}^{i-1}) )),
\end{align}
\begin{align}
   \hat{Q}, \hat{K}, \hat{V} = \mathcal{B}(G, Q), \ \mathcal{B}(G, K), \ \mathcal{B}(G, V),
\end{align}
\begin{equation}
    \hat{x}=MSA(\hat{Q}, \hat{K}, \hat{V})+x_{\mathrm{low}}^{i-1},
\end{equation}
\begin{equation}
   x_{\mathrm{low}}^{i}=FFN( LN( \hat{x}))+\hat{x},
\end{equation}
where  LN, MSA, and $\mathcal{S}$ refer to layer normalization, multi-head self-attention, and split operation, respectively.  $x_{\mathrm{low}}^{i-1}$ represents the input embeddings of the current GPTB. The BBGM $\mathcal{B}$ comprises three part—fusion, gating, and interaction—that collectively ensure a comprehensive integration of features. The BBGM can be expressed as follows:
\begin{equation}
    W_{a}, W_{b} = \mathcal{S} \left( \delta \left( \mathrm{Conv} \left( \mathrm{Conv}(G) \odot \mathrm{Conv}(F) \right) \right) \right),
\end{equation}
\begin{equation}
    \hat{F} = \mathcal{C} \left[ F + W_{a} \odot G, G + W_{a} \odot F \right],
\end{equation}
where $F$ and $\hat{F}$ represent the original embedding and the features fused with the global prior $G$. $\mathcal{C}$ and $\delta$ refer to concat operation and Gelu function. 

\noindent \textbf{Regularization}. In the first stage, we primarily constrain the zero-frequency component of the generated image, focusing on learning global representation. Given that the output of the first stage is $O_{s1}$, zero-frequency regularization $\mathcal{L}_{zf}$ can be mathematically expressed as:
\begin{equation}
    \mathcal{L}_{zf}=||\mathcal{D}(O_{s1})(0,0)-\mathcal{D}(GT)(0,0)||_1.
\end{equation}
 \vspace{-0.3cm}
\subsection{Low Frequency Restorer }
To efficiently learn content representations, the LFR restores low-frequency information within mid-resolution space. The mid-resolution features $x_{\mathrm{mid}}\in\mathbb{R}^{\frac{H}{4}\times\frac{W}{4}\times C}$  are captured via 4x downsampling. The core operator of the LFR is the residue state space block (RSSB), also known as the Mamba block \cite{guo2025mambair,DBLP:journals/corr/abs-2408-12816,DBLP:journals/corr/abs-2412-15691}, effectively capturing long-range dependencies and global features with low computational cost. The detailed structure of the RSSB is illustrated in Figure \ref{fig:5}. The RSSB can be mathematically expressed as:
\begin{equation}
    x^{\prime}=\text{VSSM}(\text{LN}(x_{\mathrm{mid}}^{i-1}))+s\cdot x_{\mathrm{mid}}^{i-1},
\end{equation}
\begin{equation}
    x_{\mathrm{mid}}^{i}=\text{PC}(\delta_{g}( \text{DC}(\text{PC}(\text{LN}(x^{\prime}))))+s'\cdot x^{\prime}),
\end{equation}
where $PC$ and $DC$ represent point-wise and depth-wise convolution, respectively. $\delta_{g}$ is the function of non-linear gate, similar
to SimpleGate,  dividing the input
along the channel dimension into two features $\mathbf{F}_{1},\mathbf{F}_{2}\in\mathbb{R}^{H\times W\times\frac{C}{2}}$. The output is then calculated by $\delta_{g}(F)=\delta(\mathbf{F}_{1}) \cdot \mathbf{F}_{2}$. The detailed description of the vision state-space module (VSSM) \cite{guo2025mambair} can be found in the supplementary materials.

\noindent \textbf{Regularization}. In this stage, our objective is to learn the low-frequency components,  with a primary focus on the restoration of coarse-grained structures and content. Given that the output of this stage is $O_{s2}$ and frequency cutoff $k$ is hyper parameter, low-frequency regularization $\mathcal{L}_{lf}$ can be mathematically expressed as:
\vspace{-0.3cm}
\begin{equation}
\begin{split}
    \mathcal{L}_{lf} = &\sum_{i=0}^{k} \sum_{j=0}^{k} \left\| \mathcal{D}(O_{s2})(i, j) - \mathcal{D}(GT)(i, j) \right\|_1, \hfill \\
    &\hfill \hspace{8.25 em} \text{where } (i,j) \neq (0,0).
\end{split}
\end{equation}


\begin{table}[t]
    \caption{Linear system means UHDformer without all non-linear functions. The difference on UHD-LL  \cite{li2023embedding} suggests that the non-linear functions mainly injects high-frequency information.}
     \vspace{-0.8em}
	\centering
	
	\resizebox{0.95\linewidth}{!}{
		\begin{tabular}{c| c c c  c |c c}
			\specialrule{1.2pt}{0.2pt}{1pt}
			Method&\multicolumn{2}{c}{Linear system} &\multicolumn{2}{c}{UHDformer} &  \multicolumn{2}{|c}{Difference~} \\
            \cmidrule(lr){1-1}
			\cmidrule(lr){2-5}
			\cmidrule(lr){6-7}
			Frequency& Low& High & Low & High & Low  & High \\
			\midrule
			PSNR $\uparrow$&28.70  &34.68& 27.92 & 36.82 & -0.78 & \textbf{2.14}\\
			SSIM $\uparrow$&0.9873& 0.8539 & 0.9820& 0.9081 & -0.0053 &\textbf{0.0542}\\
			
			\specialrule{1.2pt}{0.2pt}{1pt}
	\end{tabular}}
    \vspace{-1.3em}
	\label{tab:nonlinear}
	
\end{table}

\begin{figure*}[t]
	\centering
	\includegraphics[width=0.95\linewidth]{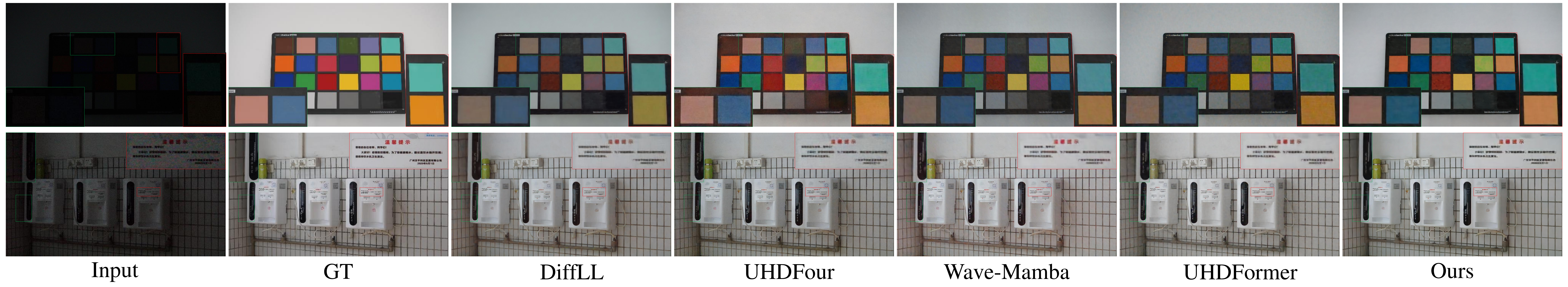}
    \vspace{-0.3cm}
	\caption{Visual comparison with other SOTA methods on the UHD-LL \cite{li2023embedding}.} 
	\vspace{-0.3cm}
	\label{fig:7}
\end{figure*}

\vspace{-0.5em}
\vspace{-0.6em}
\subsection{High Frequency  Refiner }
In the final stage, the HFR employs the designed FW-KAN to refine image details and textures within full-resolution space, further achieving high-quality image reconstruction.
\noindent \textbf{The motivation for KAN}.  Existing study \cite{deng2024exploring} on explainability in deep learning introduces an intriguing perspective: the linear system acts as a low-frequency learner, while \textit{the non-linear system injects high-frequency information}. Inspired by this, we further investigate this behavior in UHDformer by removing all non-linear activation functions. 
Table \ref{tab:nonlinear} 
validates that the non-linear function primarily injects high-frequency information. Recently, KAN \cite{liu2024kan} has become well-known as a learnable non-linear operator with powerful non-linear expression capabilities, which motivates us to explore it as the core operator in our final stage. 

\noindent \textbf{FW-KAN}.  Group-rational KAN (GR-KAN) \cite{yang2024kolmogorov} addresses the high complexity and optimization challenges of the original KAN \cite{liu2024kan} by introducing rational activation functions and variance-preserving initialization. However, applying GR-KAN in the UHD original resolution space continues to require substantial memory. Additionally, due to the pixel density of UHD images, high-frequency details become exceedingly complex. To alleviate these issues, we adopt a window partition (WP) in the DCT spectrum to partition high- and low-frequency information into blocks, which not only reduces memory consumption but also encourages the model to concentrate more effectively on high-frequency detail learning, as illustrated in the detailed FW-KAN diagram in Figure \ref{fig:3}. FW-KAN consists of multiple stacked GR-KAN. Given the original resolution feature $x_{high} \in \mathbb{R}^{H\times W\times C}$, HFR can be expressed as:
\begin{equation}
    x^{\prime}=\mathcal{D}^{-1}(\text{WR}(\text{FW-KAN}(\text{WP}(\mathcal{D}(x_{\mathrm{high}}))))),
\end{equation}
where $\mathcal{D}^{-1}$ and $\text{WR}$ stand for inverse DCT and window reverse operation.

\noindent \textbf{Regularization}. In the final stage, our objective is to inject high-frequency information, with an emphasis on the refinement of fine-grained texture and details. Given that the output of this stage is $O_{s3}$, 
high-frequency regularization $\mathcal{L}_{lf}$ can be mathematically expressed as:
\begin{equation}
\begin{split}
    \mathcal{L}_{hf} = &\sum_{i=0}^{N} \sum_{j=0}^{N} \left\| \mathcal{D}(O_{s3})(i, j) - \mathcal{D}(GT)(i, j) \right\|_1, \hfill \\
    &\hfill \hspace{6.25 em} \text{where } (i \geq k) \lor (j \geq k),
\end{split}
\end{equation}
where $N$ denotes the maximum horizontal and vertical index of the spectrum, and $k$ represents the frequency cutoff.

\begin{figure*}[t]
	\centering
	\includegraphics[width=0.96\linewidth]{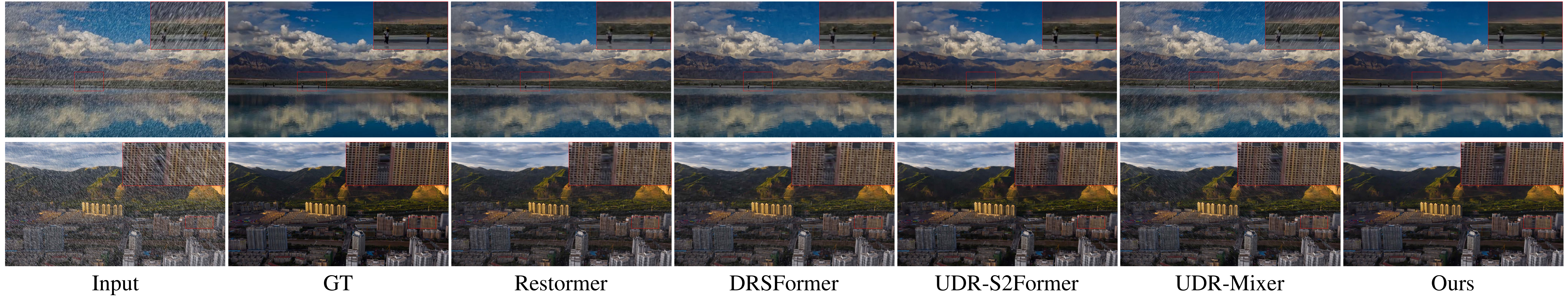}
    \vspace{-0.3cm}
	\caption{Visual comparison with other SOTA methods on the 4K-Rain13k \cite{chen2024towards}.} 
	\vspace{-0.3cm}
	\label{fig:6}
\end{figure*}

\begin{figure*}[t]
	\centering
	\includegraphics[width=0.96\linewidth]{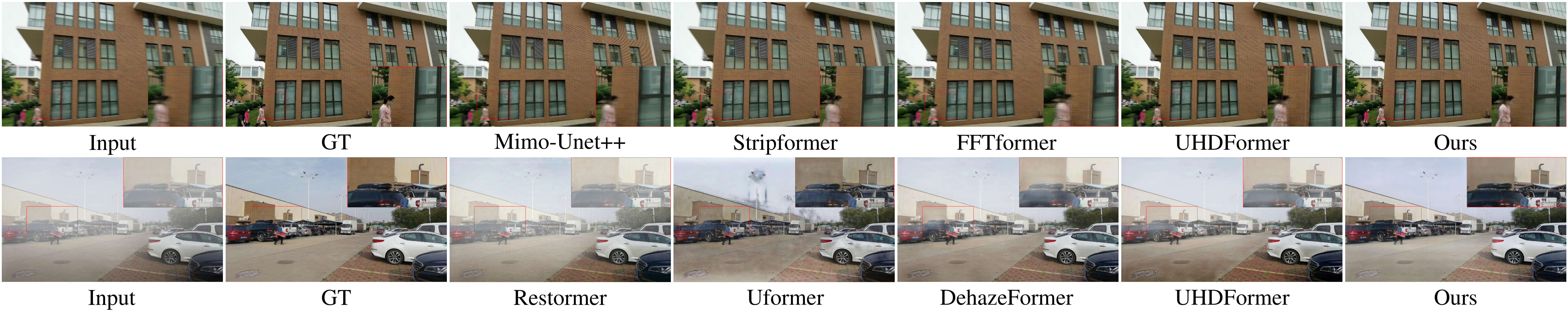}
    \vspace{-0.3cm}
	\caption{Visual comparison with other SOTA methods on the UHD-Haze and UHD-Blur \cite{wang2024correlation}.} 
	\vspace{-0.6cm}
	\label{fig:8}
\end{figure*}

\subsection{Loss }
At any stage $l$, we employ L1 and SSIM loss, and the reconstruction loss $\mathcal{L}_{rec}$ can be expressed as:
\begin{equation}
    \mathcal{L}_{rec}
=\sum_{l=1}^3\left[\mathcal{L}_{1}(O_{l},GT)+\mathcal{L}_{ssim}(O_{l},GT)\right],
\end{equation}
where $O_{l}$ represents the output at stage $l$. The total loss function $\mathcal{L}_{total}$ can be defined as: 
\begin{equation}
    \mathcal{L}_{total}
=\mathcal{L}_{rec}+\mathcal{L}_{zf}+\mathcal{L}_{lf}+\mathcal{L}_{hf}.
\end{equation}

\begin{table}[!t]
    \caption{\small{Comparison of quantitative results on UHD-LL \cite{li2023embedding}.}}
     \vspace{-0.8em}
	\vspace{-1.2em}
	\begin{center}
		\resizebox{8cm}{!}{
		\begin{tabular}{c|c|c|c|c}
			\hline
			Methods & PSNR$\uparrow$ & SSIM$\uparrow$ & LPIPS$\downarrow$   &Parameter$\downarrow$ \\
			\hline
            IFT \cite{zhao2021deep}   & 21.96 & 0.870 & 0.324 & 11.56M \\
            $\text{SNR-Aware}$  \cite{xu2022snr} & 22.72 & 0.877 & 0.304 & 40.08M \\
			Uformer \cite{wang2022uformer}   & 19.28 & 0.849 & 0.356 & 20.62M \\
            Restormer \cite{zamir2022restormer}   & 22.25 & 0.871 & 0.289 & 26.11M \\
            DiffLL \cite{jiang2023low} & 21.36 & 0.872 & 0.239 & 17.29M \\
            LLFormer \cite{wang2023ultra} & 22.79 & 0.853 & 0.264 & 13.15M \\
            UHDFour \cite{li2023embedding}  & 26.22 & 0.900 & 0.239 & 17.54M \\
			Wave-Mamba \cite{zou2024wave} & $\underline{27.35}$ & 0.913 & \textbf{0.185} & 1.258M \\
	        UHDFormer \cite{wang2024correlation} & 27.11 & $\underline{0.927}$ & 0.245 & \textbf{0.339M} \\
            \hline
			$\text{Ours}_{stage1}$  & 24.61 & 0.841 & 0.455 & - \\
            $\text{Ours}_{stage2}$  & 27.33 & 0.925 & 0.226 & - \\
            Ours  & \textbf{27.57} & \textbf{0.932} & $\underline{0.214}$ & \underline{1.131M} \\
			\hline
		\end{tabular}
		}
	\end{center}
	\label{table:ll}
	\vspace{-1.3em}
\end{table}

\begin{table}[!t]
\vspace{-0.5em}
    \caption{\small{Comparison of quantitative results on 4K-Rain13k \cite{chen2024towards}.}}
	\vspace{-1.2em}
     \vspace{-0.8em}
	\begin{center}
		\resizebox{8cm}{!}{
		\begin{tabular}{c|c|c|c|c}
			\hline
			Methods & PSNR$\uparrow$ & SSIM$\uparrow$ & LPIPS$\downarrow$ & Parameter$\downarrow$ \\
			\hline
            JORDER-E  \cite{yang2019joint} & 30.46 & 0.912 & 0.209 & 4.21M \\
            RCDNet \cite{wang2020model}   & 30.83 & 0.921 & 0.196 & 3.17M \\
            SPDNet \cite{yi2021structure}  & 31.81 & 0.922 & 0.195 & \underline{3.04M} \\
            IDT \cite{xiao2022image} & 32.91 & 0.948 & 0.124 & 16.41M \\
            Restormer \cite{zamir2022restormer}& 33.02 & 0.933 & 0.173 & 26.12M \\
            DRSformer \cite{chen2023learning} & 32.94 & 0.933 & 0.171 & 33.65M \\
            UDR-S2Former \cite{chen2023learning}& 33.36 & 0.946 & $\underline{0.122}$ & 8.53M \\
	        UDR-Mixer \cite{chen2024towards} & 34.28 & $\underline{0.951}$ & 0.133 & 4.90M \\
            \hline
			$\text{Ours}_{stage1}$  & 27.13 & 0.827 & 0.350 & - \\
            $\text{Ours}_{stage2}$  & 34.21 & 0.943 & 0.131 & -\\
            Ours  & \textbf{34.48} & \textbf{0.952} & \textbf{0.120} & \textbf{1.131M} \\
			\hline
		\end{tabular}
		}
	\end{center}
	\label{table:rain}
	\vspace{-1.3em}
    \vspace{-0.5cm}
\end{table}

\vspace{-1.3em}
\section{Experiments}
\vspace{-0.2cm}
In this section, we evaluate our approach in comparison to SOTA methods across four public UHD image restoration benchmarks: enhancement of low-light images, removal of rain artifacts, image deblurring, and haze reduction.
\vspace{-0.5em}
\subsection{Experimental Settings}

\noindent \textbf{Implementation details}. All experiments are conducted on a NVIDIA A6000 GPU. We train the models using the AdamW optimizer with an initial learning rate of 0.0005, which is gradually reduced to 1e-7 after 100k iterations using cosine annealing \cite{loshchilov2016sgdr}. The patch size is set to $512\times512$, with a batch size of 6.

\noindent \textbf{Datasets}. We evaluate our method on four UHD  restoration benchmarks. For low-light enhancement, we conduct experiments using the UHD-LL dataset \cite{li2023embedding}. The image deraining performance is assessed on the 4K-Rain13k dataset \cite{chen2024towards}. For the tasks of image dehazing and deblurring, we utilize the UHD-Haze \cite{wang2024correlation}  and UHD-Blur \cite{wang2024correlation}  datasets.

\noindent \textbf{Evaluation}. We mainly adopt peak signal to noise
ratio (PSNR) \cite{hore2010image}  and structural similarity (SSIM) \cite{wang2004image}  to evaluate the performance
of networks. Additionally, LPIPS \cite{zhang2018unreasonable} are utilized to evaluate 
perceptual performance. For methods incapable of full-resolution inference, following \cite{li2023embedding}, we adopted a splitting strategy.

\vspace{-0.6em}
\subsection{Comparisons with State-of-the-Art Methods}

\noindent \textbf{Low-light image enhancement}. We trained the proposed ERR model on the UHD-LL dataset and compared it against state-of-the-art low-light enhancement methods, including IFT \cite{zhao2021deep}, $\text{SNR-Aware}$ \cite{xu2022snr}, Uformer \cite{wang2022uformer}, Restormer \cite{zamir2022restormer}, DiffLL \cite{jiang2023low}, LLFormer \cite{wang2023ultra}, UHDFour \cite{li2023embedding}, UHDFormer \cite{wang2024correlation}, and Wave-Mamba \cite{zou2024wave}. The quantitative results in Table \ref{table:ll} demonstrate that our ERR significantly enhances performance, improving PSNR and SSIM metrics and surpassing all baselines. Figure \ref{fig:7} further provides visual evaluations on the UHD-LL dataset, where our ERR preserves finer details and achieves superior perceptual quality.


\noindent \textbf{Image deraining}. We evaluate the effectiveness of UHD deraining on the 4K-Rain13k, comparing our ERR with recent methods, including JORDER-E \cite{yang2019joint}, RCDNet \cite{wang2020model}, SPDNet \cite{yi2021structure}, IDT \cite{xiao2022image}, Restormer \cite{zamir2022restormer}, UDR-S2Former \cite{chen2023learning}, and UDR-Mixer \cite{chen2024towards}. Quantitative results in Table \ref{table:rain} demonstrate that our method achieves the highest scores across all metrics, while reducing model parameters. The visual results in Figure \ref{fig:6} show that our method effectively removes rain streaks while preserving rich texture details. 

\noindent \textbf{Image dehazing}. Table \ref{table:haze} summarizes the quantitative  results on the UHD-Haze, comparing our method with recent advanced methods such as UHD \cite{zheng2021ultra}, Restormer \cite{zamir2022restormer}, Uformer \cite{wang2022uformer}, DehazeFormer \cite{song2023vision}, and UHDFormer \cite{wang2024correlation}. Our model achieves a PSNR improvement of 2.53dB with a relatively small number of parameters, and attains the highest SSIM score. Qualitative results in  Figure \ref{fig:8} further demonstrate that ERR effectively restores clear images, whereas other methods struggle to fully remove dense haze.

\begin{table}[!t]
    \caption{\small{Comparison of quantitative results on UHD-Haze \cite{wang2024correlation}.}}
	\vspace{-1.2em}
     \vspace{-0.8em}
	\begin{center}
		\resizebox{6.5cm}{!}{
		\begin{tabular}{c|c|c|c}
			\hline
			Methods & PSNR$\uparrow$ & SSIM$\uparrow$ &Parameter$\downarrow$ \\
			\hline
            UHD  \cite{zheng2021ultra} & 18.04 & 0.811 & 34.5M \\
            Restormer \cite{zamir2022restormer}& 12.72 & 0.693 & 26.1M \\
            Uformer \cite{wang2022uformer}  & 19.83 & 0.737 & 20.6M \\
            DehazeFormer \cite{song2023vision} & 15.37 & 0.725 & 2.5M \\
	        UHDFormer \cite{wang2024correlation} & 22.59 & \underline{0.943} & \textbf{0.339M} \\
            \hline
			$\text{Ours}_{stage1}$  & 16.15 & 0.621 & - \\
            $\text{Ours}_{stage2}$  & \underline{24.67} & 0.923 & - \\
            Ours  & \textbf{25.12} & \textbf{0.950} & \underline{1.131M} \\
			\hline
		\end{tabular}
		}
	\end{center}
	\label{table:haze}
	\vspace{-1.3em}
    \vspace{-0.6cm}
\end{table}

\noindent \textbf{Image deblurring}. We evaluate the deblurring performance of our model on the UHD-Blur dataset, comparing it with recent methods including MIMO-UNet++ \cite{cho2021rethinking}, Restormer \cite{zamir2022restormer}, Uformer \cite{wang2022uformer}, Stripformer \cite{tsai2022stripformer}, FFTformer \cite{kong2023efficient}, and UHDFormer \cite{wang2024correlation}. As shown in Table \ref{table:blur}, our approach achieves the highest PSNR and SSIM scores, demonstrating its superior effectiveness. Qualitative results in Figure \ref{fig:8}  further reveal that ERR generates clearer details.

\begin{table}[!t]
    \caption{\small{Comparison of quantitative results on UHD-Blur \cite{wang2024correlation}.}}
     \vspace{-0.8em}
	\vspace{-1.2em}
	\begin{center}
		\resizebox{6.5cm}{!}{
		\begin{tabular}{c|c|c|c}
			\hline
			Methods & PSNR$\uparrow$ & SSIM$\uparrow$ &Parameter$\downarrow$ \\
			\hline
            MIMO-Unet++ \cite{cho2021rethinking}   & 25.03 & 0.811 & 16.1M \\
            Restormer \cite{zamir2022restormer}& 25.21 & 0.693 & 26.1M \\
            Uformer \cite{wang2022uformer} & 25.27 & 0.737 & 20.6M \\
            Stripformer \cite{tsai2022stripformer} & 25.05 & 0.725 & 19.7M \\
            FFTformer \cite{kong2023efficient}& 25.41 & 0.725 & 16.6M \\
	        UHDFormer & 28.82 & 0.844 & \textbf{0.339M} \\
            \hline
			$\text{Ours}_{stage1}$  & 24.16 & 0.705 & - \\
            $\text{Ours}_{stage2}$  & \underline{29.68} & \underline{0.857} & - \\
            Ours  & \textbf{29.72} & \textbf{0.861} & \underline{1.131M} \\
			\hline
		\end{tabular}
		}
	\end{center}
	\label{table:blur}
	\vspace{-1.3em}
\end{table}

\vspace{-0.6em}
\subsection{Ablation Study}
\vspace{-0.5em}
We perform comprehensive ablation experiments to verify the effectiveness of each contribution and design. All ablations are conducted on UHD-LL \cite{li2023embedding}.

\noindent \textbf{Effect of different frequency cutoffs $k$}.
   The ablation results with different cutoffs $k$ are shown in Table \ref{tab:cutoff}, which indicate that both excessively large and small values of  $k$ negatively affect performance. A small  $k$ overcomplicates high-frequency component, increasing the learning difficulty of the HFR, while a large  $k$ introduces excessive low-frequency component, hindering the restoration of the LFR. 
   
\begin{table}[t]
    \vspace{-0.2cm}
    \caption{ Effect of different frequency cutoffs $k$.}
     \vspace{-1em}
	\centering

	\resizebox{0.8\linewidth}{!}{
  {\setlength{\tabcolsep}{3pt}
		\begin{tabular}{c| c c c  c c }
			\specialrule{1.2pt}{0.2pt}{1pt}
			The value of $k$ & 32& 64 &96 & 128 & 256 \\
			\midrule
			PSNR $\uparrow$&  26.71    &     \textbf{27.57} &  27.52    &  27.29    &   27.25   \\
			SSIM $\uparrow$&  0.9296  & \textbf{0.9326}  &   0.9323  &0.9319    &    0.9313 \\
			
			\specialrule{1.2pt}{0.2pt}{1pt}
	\end{tabular}}
    }
    \vspace{-1.8em}
	\label{tab:cutoff}
	
\end{table}

\begin{table}[h!]
\caption{Ablation study with different architecture. PR refers to progressive residual, as indicated by the yellow arrows between each stage in Figure \ref{fig:3}.}
 \vspace{-0.8em}
\centering
\label{tab:freq}

\setlength{\tabcolsep}{3pt} 
\renewcommand{\arraystretch}{1} 
\footnotesize 

\begin{tabular}{c|ccccccc|cc}
\specialrule{1.2pt}{0.2pt}{1pt}

Method  & $\mathcal{L}_{zf}$ & $\mathcal{L}_{lf}$  & $\mathcal{L}_{hf}$  & ZFE & LFR & HFR & PR & PSNR$\uparrow$ & SSIM$\uparrow$       \\\midrule
\midrule
A &  $\times$ & $\checkmark$ & $\checkmark$ & $\checkmark$ & $\checkmark$& $\checkmark$  & $\checkmark$  & 26.73 & 0.9286 \\
B &  $\checkmark$&  $\times$ & $\checkmark$ & $\checkmark$ & $\checkmark$& $\checkmark$  & $\checkmark$  & 26.57 & 0.9277 \\
C & $\checkmark$& $\checkmark$ & $\times$ & $\checkmark$ & $\checkmark$& $\checkmark$  & $\checkmark$ & 26.80  & 0.9290\\
\midrule
D &  $\times$ & $\checkmark$ & $\checkmark$ & $\times$ & $\checkmark$& $\checkmark$  & $\checkmark$ &27.01  & 0.9311 \\
E & $\checkmark$&  $\times$ & $\checkmark$ & $\checkmark$ & $\times$ & $\checkmark$  & $\checkmark$  & 26.81 & 0.9186 \\ 
F & $\checkmark$& $\checkmark$ & $\times$ & $\checkmark$ & $\checkmark$& $\times$   & $\checkmark$ & 27.04  & 0.9266 \\
\midrule
G & $\checkmark$& $\checkmark$ & $\checkmark$ & $\checkmark$ & $\checkmark$& $\checkmark$   & $\times$ &  26.99 & 0.9307
  \\
\midrule
\midrule
H & $\checkmark$& $\checkmark$ & $\checkmark$ & $\checkmark$ & $\checkmark$&$\checkmark$&$\checkmark$ & \textbf{27.57} & \textbf{0.9326 } \\
\specialrule{1.2pt}{0.2pt}{1pt}

\end{tabular}
 \vspace{-0.5em}
\label{tab:framerwork}
\end{table}

\begin{table}[t]
	\caption{Ablation study with the detail design of the ZFE. }
	\centering
	 \vspace{-1.2em}
	\resizebox{1\linewidth}{!}{
		\begin{tabular}{c c c  c c |c c}
			\specialrule{1.2pt}{0.2pt}{1pt}
			\multicolumn{2}{c}{AAP} &\multicolumn{3}{c}{BBGM} &  \multicolumn{2}{c}{Metrics~} \\
			\cmidrule(lr){1-2}
			\cmidrule(lr){3-5}
            \cmidrule(lr){6-7}
			 $AvP_{1,1}$ & $AvP_{\frac{H}{8}, \frac{W}{8}}(x)$ & Fusion & Gate & Interaction & PSNR$\uparrow$ & SSIM$\uparrow$ \\
			\midrule
			\midrule
			$\times$  &$\times$ & $\times$ &$\times$ &$\times$ & 26.70  & 0.9289\\
			$\times$& $\checkmark$ & $\checkmark$& $\checkmark$ &$\checkmark$ & 27.16 & 0.9315\\
            $\checkmark$& $\times$ & $\checkmark$& $\checkmark$ &$\checkmark$ & 27.18 & 0.9316  \\
            \midrule

			$\checkmark$ & $\checkmark$ & $\times$& $\checkmark$   & $\checkmark$ & 26.00& 0.9251 \\
			$\checkmark$ & $\checkmark$ & $\checkmark$ & $\times$&$\checkmark$ & 26.20 &0.9268 \\
            $\checkmark$ & $\checkmark$ & $\checkmark$ & $\checkmark$&$\times$ & 26.04 &0.9256 \\
			\midrule
            \midrule
			$\checkmark$ & $\checkmark$ & $\checkmark$ &  $\checkmark$&$\checkmark$ & \textbf{27.57} & \textbf{0.9326 }\\
			\specialrule{1.2pt}{0.2pt}{1pt}
	\end{tabular}}
	\label{tab:zfe}
	 \vspace{-0.8 em}
\end{table}

\noindent \textbf{Ablation with different architecture}. Table \ref{tab:framerwork} demonstrates the impact of all components on our framework. A, B, and C validate the effectiveness of our frequency regularizations, while D, E, and F confirm the efficacy of each stage's network. The superior performance of D, E, and F over A, B, and C suggests that, without the frequency constraints, the networks face challenges in optimization. H outperforms G, indicating that the progressive residual is more effective than using the input at each stage.

\begin{figure}[t]
	\centering
	\includegraphics[width=0.98\linewidth]{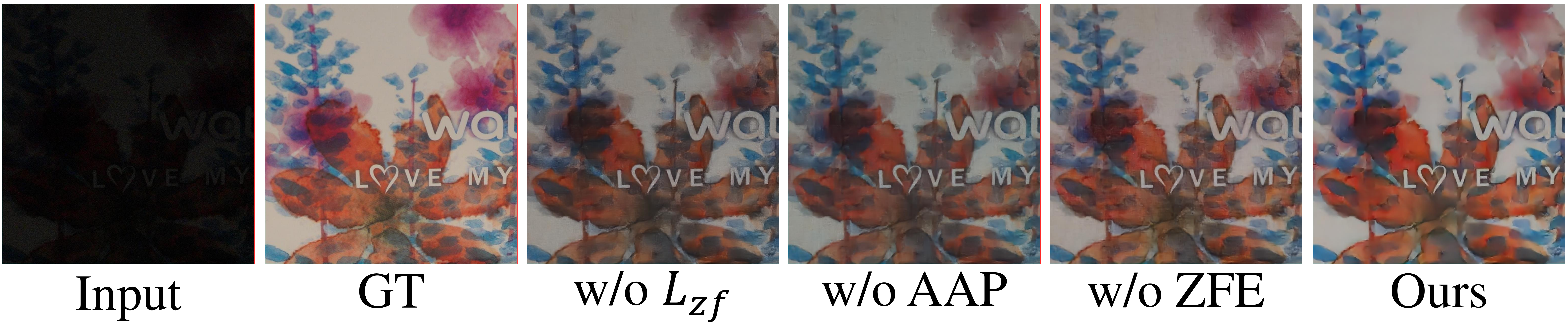}
    \vspace{-0.4 cm}
	\caption{Visual ablation results for the zero-frequency part. } 
	\vspace{-0.3 cm}
	\label{fig:ablow}
\end{figure}

\noindent \textbf{Ablation with the detail design of the ZFE}. In Table \ref{tab:zfe}, we discuss the details of the ZFE, focusing on the AAP unit for global prior generation and the BBGM for global prior integration.  First, we remove the entire AAP unit, as well as the global and local $AvP$ separately, and this results confirm the effectiveness of our prior design. Next, removing each module in the BBGM leads to poorer performance, indicating the critical role of prior integration strategy for model performance. Figure \ref{fig:ablow} illustrates visual ablation results for the zero-frequency part.

\begin{figure}[t]
	\centering
	\includegraphics[width=0.98\linewidth]{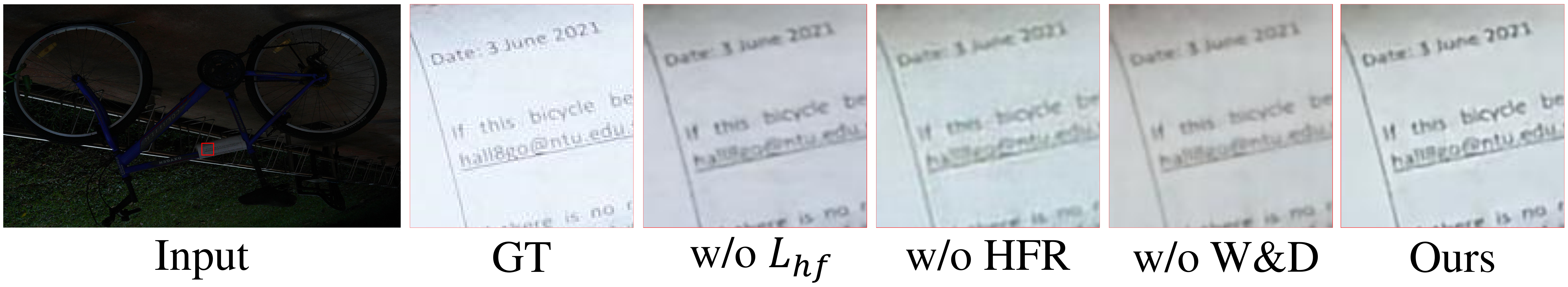}

        \vspace{-0.3 cm}
	\caption{Visual ablation results for the high-frequency part. } 
	\vspace{-0.3 cm}
	\label{fig:abhigh}
\end{figure}

\noindent \textbf{Ablation with the detail design of the HFR}.  Table \ref{tab:hfr} validates the effectiveness of DCT and WR operations in HFR. The models without DCT or WR achieve lower performance, suggesting that window partitioning in the DCT spectrum enhances the model's ability to capture high-frequency information more effectively. The model without WP \& DCT (W \& D) achieves the lowest performance, indicating that WP and DCT alone are effective in improving performance. Figure \ref{fig:abhigh} illustrates visual ablation results for the high-frequency part, where our method achieves the best visual results in terms of details and textures.

\begin{table}[t]
    \caption{ Ablation study with the detail design of the HFR.}
	\centering
 \vspace{-0.6em}
	\resizebox{0.9\linewidth}{!}{
  {\setlength{\tabcolsep}{3pt}
		\begin{tabular}{c| c c c | c  }
			\specialrule{1.2pt}{0.2pt}{1pt}
			Model & w/o DCT &  w/o WP&  w/o WP \& DCT  & Full Model \\
			\midrule
			PSNR $\uparrow$&  26.88    &   26.97  &   25.78  &   \textbf{27.57}   \\
			SSIM $\uparrow$&   0.9308  &    0.9296   &  0.9247 & \textbf{0.9326 }     \\
			
			\specialrule{1.2pt}{0.2pt}{1pt}
	\end{tabular}}
    }
    \vspace{-1.0em}
	\label{tab:hfr}
	
\end{table}

\noindent \textbf{Ablation for FW-KAN}. In Table \ref{tab:kan}, we compare our FW-KAN with other nonlinear operators, specifically MLP and the original KAN. For MLP, we employ a two-layer activation structure, defined as linear-ReLU-linear-ReLU-linear, to enhance nonlinear expressiveness. MLP-6, MLP-12, and MLP-24 refer to MLPs with 6, 12, and 24 layers. Experiments show that, compared to MLP, our FW-KAN can learn complex representations with fewer parameters, while the original KAN encounters optimization challenges.

\noindent \textbf{Inference efficiency}. Table \ref{tab:time} presents a comparison of inference times between our method and  baselines. LLFormer, which cannot perform full-resolution inference, incurs a high time cost. 
Our approach demonstrates the fastest inference efficiency among recent UHD methods.

\begin{table}[t]
    \caption{ Ablation for FW-KAN.}
        \vspace{-0.6em}
	\centering

	\resizebox{1\linewidth}{!}{
  {\setlength{\tabcolsep}{4pt}
		\begin{tabular}{c| c c c c | c  }
			\specialrule{1.2pt}{0.2pt}{1pt}
			Model & MLP-6 &  MLP-12 &  MLP-24 &  KAN  & FW-KAN \\
			\midrule
			PSNR $\uparrow$&  26.03 &   26.58  &  26.97 & 20.48  &    \textbf{27.57}   \\
			SSIM $\uparrow$&   0.9294 &    0.9297   &0.9311  & 0.8836  &  \textbf{0.9326 }     \\
            Parameter$\downarrow$ &   30.73K   &    35.63K & 45.42K & 27.57K &  \textbf{27.47}K   \\
			
			\specialrule{1.2pt}{0.2pt}{1pt}
	\end{tabular}}
    }
    \vspace{-1.1em}
	\label{tab:kan}
	
\end{table}

\begin{table}[t]
    \caption{Comparison of inference time on 4K images.}
        \vspace{-0.8em}
	\centering

	\resizebox{1\linewidth}{!}{
  {\setlength{\tabcolsep}{3pt}
		\begin{tabular}{c| c c c  | c  }
			\specialrule{1.2pt}{0.2pt}{1pt}
			Model&LLformer& UHDformer  &  Wave-Mamba  & Ours \\
			\midrule
			Inference time (s) $\downarrow$&    41.056   &   0.887     &  0.618  &       \textbf{0.532}   \\

			\specialrule{1.2pt}{0.2pt}{1pt}
	\end{tabular}}
    }
    \vspace{-1 em}
	\label{tab:time}
	
\end{table}

\begin{figure}[t]
	\centering
	\includegraphics[width=0.98\linewidth]{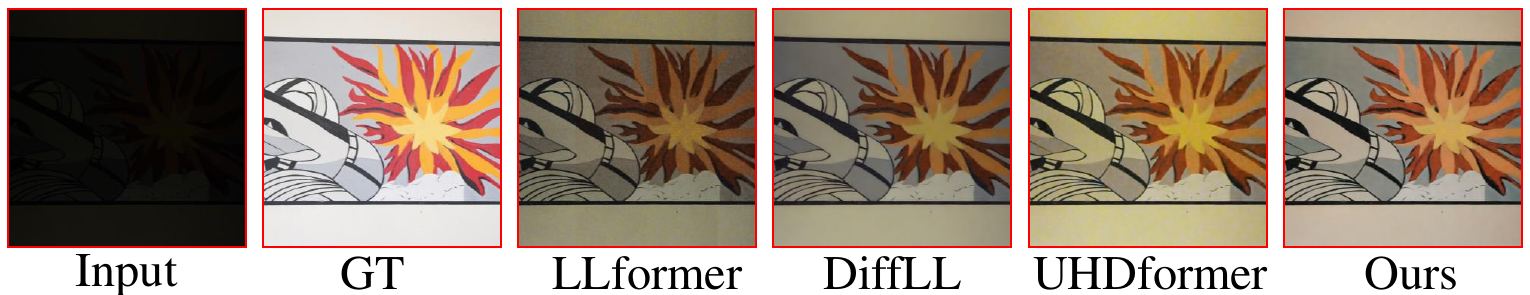}
    \vspace{-0.2 cm}
	\caption{Limitations in extremely low-light scenarios. } 
	\vspace{-0.6 cm}
	\label{fig:limi}
\end{figure}

\vspace{-0.4 cm}
\section{Conclusion}
\vspace{-0.2 cm}
In the paper, we develop a novel framework, namely
ERR,  which comprises three sub-networks: the ZFE to capture global information, the LFR to reconstruct primary content, and the HFR 
for detail refinement.  ERR shows
SOTA performance on multiple UHD tasks, and extensive ablation experiments prove that each component is effective.

\noindent \textbf{Limitations and future works.} Although our method achieves outstanding performance, it remains challenging to achieve satisfactory results in extreme scenarios, as shown in Figure \ref{fig:limi}. Therefore, mitigating this weakness will be the focus of our future research.

\noindent \textbf{Acknowledgments.} This work was supported by Natural Science Foundation of China: No. 62406135, Natural Science Foundation of Jiangsu Province: BK20241198, Gusu Innovation and Entrepreneur Leading Talents: No. ZXL2024362 and Nanjing University-China Mobile Communications Group Co. Ltd. Joint Institute.

{
    \small
    \bibliographystyle{ieeenat_fullname}
    \bibliography{main}

\begin{thebibliography}{95}
\providecommand{\natexlab}[1]{#1}
\providecommand{\url}[1]{\texttt{#1}}
\expandafter\ifx\csname urlstyle\endcsname\relax
  \providecommand{\doi}[1]{doi: #1}\else
  \providecommand{\doi}{doi: \begingroup \urlstyle{rm}\Url}\fi

\bibitem[Chen et~al.(2024{\natexlab{a}})Chen, Chen, Wu, Zheng, Pan, and Fu]{chen2024towards}
Hongming Chen, Xiang Chen, Chen Wu, Zhuoran Zheng, Jinshan Pan, and Xianping Fu.
\newblock Towards ultra-high-definition image deraining: A benchmark and an efficient method.
\newblock \emph{arXiv preprint arXiv:2405.17074}, 2024{\natexlab{a}}.

\bibitem[Chen et~al.(2023)Chen, Li, Li, and Pan]{chen2023learning}
Xiang Chen, Hao Li, Mingqiang Li, and Jinshan Pan.
\newblock Learning a sparse transformer network for effective image deraining.
\newblock In \emph{Proceedings of the IEEE/CVF Conference on Computer Vision and Pattern Recognition}, pages 5896--5905, 2023.

\bibitem[Chen et~al.(2024{\natexlab{b}})Chen, Pan, and Dong]{chen2024bidirectional}
Xiang Chen, Jinshan Pan, and Jiangxin Dong.
\newblock Bidirectional multi-scale implicit neural representations for image deraining.
\newblock In \emph{Proceedings of the IEEE/CVF Conference on Computer Vision and Pattern Recognition}, pages 25627--25636, 2024{\natexlab{b}}.

\bibitem[Chen et~al.(2024{\natexlab{c}})Chen, Li, Wang, Chen, Jiang, Li, Wang, Yang, and Tai]{DBLP:journals/corr/abs-2411-06558}
Zhennan Chen, Yajie Li, Haofan Wang, Zhibo Chen, Zhengkai Jiang, Jun Li, Qian Wang, Jian Yang, and Ying Tai.
\newblock Region-aware text-to-image generation via hard binding and soft refinement.
\newblock \emph{CoRR}, abs/2411.06558, 2024{\natexlab{c}}.

\bibitem[Cheng and Sun(2024)]{cheng2024spt}
Senlin Cheng and Haopeng Sun.
\newblock Spt: Sequence prompt transformer for interactive image segmentation.
\newblock \emph{arXiv preprint arXiv:2412.10224}, 2024.

\bibitem[Cho et~al.(2021)Cho, Ji, Hong, Jung, and Ko]{cho2021rethinking}
Sung-Jin Cho, Seo-Won Ji, Jun-Pyo Hong, Seung-Won Jung, and Sung-Jea Ko.
\newblock Rethinking coarse-to-fine approach in single image deblurring.
\newblock In \emph{Proceedings of the IEEE/CVF international conference on computer vision}, pages 4641--4650, 2021.

\bibitem[Cui et~al.(2023)Cui, Ren, Cao, and Knoll]{cui2023image}
Yuning Cui, Wenqi Ren, Xiaochun Cao, and Alois Knoll.
\newblock Image restoration via frequency selection.
\newblock \emph{IEEE Transactions on Pattern Analysis and Machine Intelligence}, 2023.

\bibitem[Deng et~al.(2024)Deng, Xu, Duan, Wu, Shu, and Deng]{deng2024exploring}
Haoyu Deng, Zijing Xu, Yule Duan, Xiao Wu, Wenjie Shu, and Liang-Jian Deng.
\newblock Exploring the low-pass filtering behavior in image super-resolution.
\newblock \emph{arXiv preprint arXiv:2405.07919}, 2024.

\bibitem[Deng et~al.(2021)Deng, Ren, Yan, Wang, Song, and Cao]{deng2021multi}
Senyou Deng, Wenqi Ren, Yanyang Yan, Tao Wang, Fenglong Song, and Xiaochun Cao.
\newblock Multi-scale separable network for ultra-high-definition video deblurring.
\newblock In \emph{Proceedings of the IEEE/CVF International Conference on Computer Vision}, pages 14030--14039, 2021.

\bibitem[Dong et~al.(2024)Dong, Zhao, Cai, and Yang]{DBLP:journals/corr/abs-2408-12816}
Chenyu Dong, Chen Zhao, Weiling Cai, and Bo Yang.
\newblock O-mamba: O-shape state-space model for underwater image enhancement.
\newblock \emph{CoRR}, abs/2408.12816, 2024.

\bibitem[Feng et~al.(2024)Feng, Li, and Loy]{DBLP:journals/corr/abs-2408-05205}
Ruicheng Feng, Chongyi Li, and Chen~Change Loy.
\newblock Kalman-inspired feature propagation for video face super-resolution.
\newblock \emph{CoRR}, abs/2408.05205, 2024.

\bibitem[Fritsche et~al.(2019)Fritsche, Gu, and Timofte]{fritsche2019frequency}
Manuel Fritsche, Shuhang Gu, and Radu Timofte.
\newblock Frequency separation for real-world super-resolution.
\newblock In \emph{2019 IEEE/CVF International Conference on Computer Vision Workshop (ICCVW)}, pages 3599--3608. IEEE, 2019.

\bibitem[Guo et~al.(2025)Guo, Li, Dai, Ouyang, Ren, and Xia]{guo2025mambair}
Hang Guo, Jinmin Li, Tao Dai, Zhihao Ouyang, Xudong Ren, and Shu-Tao Xia.
\newblock Mambair: A simple baseline for image restoration with state-space model.
\newblock In \emph{European Conference on Computer Vision}, pages 222--241. Springer, 2025.

\bibitem[Hore and Ziou(2010)]{hore2010image}
Alain Hore and Djemel Ziou.
\newblock Image quality metrics: Psnr vs. ssim.
\newblock In \emph{2010 20th international conference on pattern recognition}, pages 2366--2369. IEEE, 2010.

\bibitem[Hu et~al.(2024)Hu, Tai, Zhao, Zhao, Zhang, Li, Zhong, and Yang]{DBLP:journals/corr/abs-2412-15691}
Xiantao Hu, Ying Tai, Xu Zhao, Chen Zhao, Zhenyu Zhang, Jun Li, Bineng Zhong, and Jian Yang.
\newblock Exploiting multimodal spatial-temporal patterns for video object tracking.
\newblock \emph{CoRR}, abs/2412.15691, 2024.

\bibitem[Jiang et~al.(2023{\natexlab{a}})Jiang, Luo, Fan, Han, and Liu]{jiang2023low}
Hai Jiang, Ao Luo, Haoqiang Fan, Songchen Han, and Shuaicheng Liu.
\newblock Low-light image enhancement with wavelet-based diffusion models.
\newblock \emph{ACM Transactions on Graphics (TOG)}, 42\penalty0 (6):\penalty0 1--14, 2023{\natexlab{a}}.

\bibitem[Jiang et~al.(2023{\natexlab{b}})Jiang, Liu, Wang, Zhong, Jiang, and Lin]{jiang2023dawn}
Kui Jiang, Wenxuan Liu, Zheng Wang, Xian Zhong, Junjun Jiang, and Chia-Wen Lin.
\newblock Dawn: Direction-aware attention wavelet network for image deraining.
\newblock In \emph{Proceedings of the 31st ACM international conference on multimedia}, pages 7065--7074, 2023{\natexlab{b}}.

\bibitem[Jin et~al.(2024)Jin, Li, Wang, Zhang, and Zhang]{DBLP:conf/eccv/JinLWZZ24}
Yeying Jin, Xin Li, Jiadong Wang, Yan Zhang, and Malu Zhang.
\newblock Raindrop clarity: {A} dual-focused dataset for day and night raindrop removal.
\newblock In \emph{Computer Vision - {ECCV} 2024 - 18th European Conference, Milan, Italy, September 29-October 4, 2024, Proceedings, Part {VI}}, pages 1--17. Springer, 2024.

\bibitem[Kong et~al.(2023)Kong, Dong, Ge, Li, and Pan]{kong2023efficient}
Lingshun Kong, Jiangxin Dong, Jianjun Ge, Mingqiang Li, and Jinshan Pan.
\newblock Efficient frequency domain-based transformers for high-quality image deblurring.
\newblock In \emph{Proceedings of the IEEE/CVF Conference on Computer Vision and Pattern Recognition}, pages 5886--5895, 2023.

\bibitem[Li et~al.(2024)Li, Li, Zhu, Jin, Feng, Zhang, and Chen]{DBLP:conf/cvpr/LiLZJF0024}
Bingchen Li, Xin Li, Hanxin Zhu, Yeying Jin, Ruoyu Feng, Zhizheng Zhang, and Zhibo Chen.
\newblock Sed: Semantic-aware discriminator for image super-resolution.
\newblock In \emph{{IEEE/CVF} Conference on Computer Vision and Pattern Recognition, {CVPR} 2024, Seattle, WA, USA, June 16-22, 2024}, pages 25784--25795. {IEEE}, 2024.

\bibitem[Li et~al.(2023{\natexlab{a}})Li, Guo, Zhou, Liang, Zhou, Feng, and Loy]{li2023embedding}
Chongyi Li, Chun-Le Guo, Man Zhou, Zhexin Liang, Shangchen Zhou, Ruicheng Feng, and Chen~Change Loy.
\newblock Embedding fourier for ultra-high-definition low-light image enhancement.
\newblock \emph{arXiv preprint arXiv:2302.11831}, 2023{\natexlab{a}}.

\bibitem[Li et~al.(2023{\natexlab{b}})Li, Li, Guo, and Guo]{li2023uhdnerf}
Quewei Li, Feichao Li, Jie Guo, and Yanwen Guo.
\newblock Uhdnerf: Ultra-high-definition neural radiance fields.
\newblock In \emph{Proceedings of the IEEE/CVF International Conference on Computer Vision}, pages 23097--23108, 2023{\natexlab{b}}.

\bibitem[Li et~al.(2023{\natexlab{c}})Li, Fan, Xiang, Demandolx, Ranjan, Timofte, and Van~Gool]{li2023efficient}
Yawei Li, Yuchen Fan, Xiaoyu Xiang, Denis Demandolx, Rakesh Ranjan, Radu Timofte, and Luc Van~Gool.
\newblock Efficient and explicit modelling of image hierarchies for image restoration.
\newblock In \emph{Proceedings of the IEEE/CVF Conference on Computer Vision and Pattern Recognition}, pages 18278--18289, 2023{\natexlab{c}}.

\bibitem[Liang et~al.(2021)Liang, Cao, Sun, Zhang, Van~Gool, and Timofte]{liang2021swinir}
Jingyun Liang, Jiezhang Cao, Guolei Sun, Kai Zhang, Luc Van~Gool, and Radu Timofte.
\newblock Swinir: Image restoration using swin transformer.
\newblock In \emph{Proceedings of the IEEE/CVF international conference on computer vision}, pages 1833--1844, 2021.

\bibitem[Liu et~al.(2024)Liu, Wang, Vaidya, Ruehle, Halverson, Solja{\v{c}}i{\'c}, Hou, and Tegmark]{liu2024kan}
Ziming Liu, Yixuan Wang, Sachin Vaidya, Fabian Ruehle, James Halverson, Marin Solja{\v{c}}i{\'c}, Thomas~Y Hou, and Max Tegmark.
\newblock Kan: Kolmogorov-arnold networks.
\newblock \emph{arXiv preprint arXiv:2404.19756}, 2024.

\bibitem[Loshchilov and Hutter(2016)]{loshchilov2016sgdr}
Ilya Loshchilov and Frank Hutter.
\newblock Sgdr: Stochastic gradient descent with warm restarts.
\newblock \emph{arXiv preprint arXiv:1608.03983}, 2016.

\bibitem[Lou et~al.(2024{\natexlab{a}})Lou, Xu, Zhang, Zhang, Cao, Wang, and Huang]{DBLP:conf/bibm/LouXZZCW024}
Yiwei Lou, Dexuan Xu, Rongchao Zhang, Jiayu Zhang, Yongzhi Cao, Hanpin Wang, and Yu Huang.
\newblock {MR} image quality assessment via enhanced mamba: {A} hybrid spatial-frequency approach.
\newblock In \emph{{IEEE} International Conference on Bioinformatics and Biomedicine, {BIBM} 2024, Lisbon, Portugal, December 3-6, 2024}, pages 3561--3564. {IEEE}, 2024{\natexlab{a}}.

\bibitem[Lou et~al.(2024{\natexlab{b}})Lou, Zhang, Xu, Cao, Wang, and Huang]{DBLP:conf/icmcs/LouZXCW024}
Yiwei Lou, Jiayu Zhang, Dexuan Xu, Yongzhi Cao, Hanpin Wang, and Yu Huang.
\newblock No-reference {MRI} quality assessment via contrastive representation: Spatial and frequency domain perspectives.
\newblock In \emph{{IEEE} International Conference on Multimedia and Expo, {ICME} 2024, Niagara Falls, ON, Canada, July 15-19, 2024}, pages 1--6. {IEEE}, 2024{\natexlab{b}}.

\bibitem[Lu et~al.(2023)Lu, Liu, and Kong]{DBLP:conf/iccv/LuLK23}
Shilin Lu, Yanzhu Liu, and Adams~Wai{-}Kin Kong.
\newblock {TF-ICON:} diffusion-based training-free cross-domain image composition.
\newblock In \emph{{IEEE/CVF} International Conference on Computer Vision, {ICCV} 2023, Paris, France, October 1-6, 2023}, pages 2294--2305. {IEEE}, 2023.

\bibitem[Lu et~al.(2024{\natexlab{a}})Lu, Wang, Li, Liu, and Kong]{DBLP:conf/cvpr/LuWLLK24}
Shilin Lu, Zilan Wang, Leyang Li, Yanzhu Liu, and Adams~Wai{-}Kin Kong.
\newblock {MACE:} mass concept erasure in diffusion models.
\newblock In \emph{{IEEE/CVF} Conference on Computer Vision and Pattern Recognition, {CVPR} 2024, Seattle, WA, USA, June 16-22, 2024}, pages 6430--6440. {IEEE}, 2024{\natexlab{a}}.

\bibitem[Lu et~al.(2024{\natexlab{b}})Lu, Zhou, Lu, Zhu, and Kong]{DBLP:journals/corr/abs-2410-18775}
Shilin Lu, Zihan Zhou, Jiayou Lu, Yuanzhi Zhu, and Adams~Wai{-}Kin Kong.
\newblock Robust watermarking using generative priors against image editing: From benchmarking to advances.
\newblock \emph{CoRR}, abs/2410.18775, 2024{\natexlab{b}}.

\bibitem[Lv et~al.(2024)Lv, Zhang, Wang, Zheng, Zhong, Li, and Nie]{lv2024fourier}
Xiaoqian Lv, Shengping Zhang, Chenyang Wang, Yichen Zheng, Bineng Zhong, Chongyi Li, and Liqiang Nie.
\newblock Fourier priors-guided diffusion for zero-shot joint low-light enhancement and deblurring.
\newblock In \emph{Proceedings of the IEEE/CVF Conference on Computer Vision and Pattern Recognition}, pages 25378--25388, 2024.

\bibitem[Mao et~al.(2023)Mao, Liu, Liu, Li, Shen, and Wang]{mao2023intriguing}
Xintian Mao, Yiming Liu, Fengze Liu, Qingli Li, Wei Shen, and Yan Wang.
\newblock Intriguing findings of frequency selection for image deblurring.
\newblock In \emph{Proceedings of the AAAI Conference on Artificial Intelligence}, pages 1905--1913, 2023.

\bibitem[Mao et~al.(2024)Mao, Wang, Xie, Li, and Wang]{mao2024loformer}
Xintian Mao, Jiansheng Wang, Xingran Xie, Qingli Li, and Yan Wang.
\newblock Loformer: Local frequency transformer for image deblurring.
\newblock In \emph{Proceedings of the 32nd ACM International Conference on Multimedia}, pages 10382--10391, 2024.

\bibitem[Qin et~al.(2024)Qin, Wu, Liu, Lin, Guo, Park, and Li]{qin2024restore}
Chu-Jie Qin, Rui-Qi Wu, Zikun Liu, Xin Lin, Chun-Le Guo, Hyun~Hee Park, and Chongyi Li.
\newblock Restore anything with masks: Leveraging mask image modeling for blind all-in-one image restoration.
\newblock \emph{arXiv preprint arXiv:2409.19403}, 2024.

\bibitem[Qin et~al.(2021)Qin, Zhang, Wu, and Li]{qin2021fcanet}
Zequn Qin, Pengyi Zhang, Fei Wu, and Xi Li.
\newblock Fcanet: Frequency channel attention networks.
\newblock In \emph{Proceedings of the IEEE/CVF international conference on computer vision}, pages 783--792, 2021.

\bibitem[Shao et~al.(2023)Shao, Xu, Wen, Gao, Yang, and Huang]{DBLP:conf/iccv/ShaoXWG0H23}
Huiyang Shao, Qianqian Xu, Peisong Wen, Peifeng Gao, Zhiyong Yang, and Qingming Huang.
\newblock Building bridge across the time: Disruption and restoration of murals in the wild.
\newblock In \emph{{IEEE/CVF} International Conference on Computer Vision, {ICCV} 2023, Paris, France, October 1-6, 2023}, pages 20202--20212. {IEEE}, 2023.

\bibitem[Song et~al.(2023)Song, He, Qian, and Du]{song2023vision}
Yuda Song, Zhuqing He, Hui Qian, and Xin Du.
\newblock Vision transformers for single image dehazing.
\newblock \emph{IEEE Transactions on Image Processing}, 32:\penalty0 1927--1941, 2023.

\bibitem[Sun(2024)]{sun2024ultra}
Haopeng Sun.
\newblock Ultra-high resolution segmentation via boundary-enhanced patch-merging transformer.
\newblock \emph{arXiv preprint arXiv:2412.10181}, 2024.

\bibitem[Sun et~al.(2024{\natexlab{a}})Sun, Xu, Jin, Luo, Qian, and Liu]{sunprogram}
Haopeng Sun, Lumin Xu, Sheng Jin, Ping Luo, Chen Qian, and Wentao Liu.
\newblock Program: Prototype graph model based pseudo-label learning for test-time adaptation.
\newblock In \emph{The Twelfth International Conference on Learning Representations}, 2024{\natexlab{a}}.

\bibitem[Sun et~al.(2024{\natexlab{b}})Sun, Ren, Gao, Wang, and Cao]{DBLP:conf/eccv/SunRGWC24}
Shangquan Sun, Wenqi Ren, Xinwei Gao, Rui Wang, and Xiaochun Cao.
\newblock Restoring images in adverse weather conditions via histogram transformer.
\newblock In \emph{Computer Vision - {ECCV} 2024 - 18th European Conference, Milan, Italy, September 29-October 4, 2024, Proceedings, Part {XXII}}, pages 111--129. Springer, 2024{\natexlab{b}}.

\bibitem[Suvorov et~al.(2022)Suvorov, Logacheva, Mashikhin, Remizova, Ashukha, Silvestrov, Kong, Goka, Park, and Lempitsky]{suvorov2022resolution}
Roman Suvorov, Elizaveta Logacheva, Anton Mashikhin, Anastasia Remizova, Arsenii Ashukha, Aleksei Silvestrov, Naejin Kong, Harshith Goka, Kiwoong Park, and Victor Lempitsky.
\newblock Resolution-robust large mask inpainting with fourier convolutions.
\newblock In \emph{Proceedings of the IEEE/CVF winter conference on applications of computer vision}, pages 2149--2159, 2022.

\bibitem[Tai et~al.(2017)Tai, Yang, Liu, and Xu]{tai2017memnet}
Ying Tai, Jian Yang, Xiaoming Liu, and Chunyan Xu.
\newblock Memnet: A persistent memory network for image restoration.
\newblock In \emph{Proceedings of the IEEE international conference on computer vision}, pages 4539--4547, 2017.

\bibitem[Tian et~al.(2022)Tian, Yan, Zhai, Guo, and Gao]{DBLP:journals/ijcv/TianYZGG22}
Yuan Tian, Yichao Yan, Guangtao Zhai, Guodong Guo, and Zhiyong Gao.
\newblock {EAN:} event adaptive network for enhanced action recognition.
\newblock \emph{Int. J. Comput. Vis.}, 130\penalty0 (10):\penalty0 2453--2471, 2022.

\bibitem[Tian et~al.(2023{\natexlab{a}})Tian, Lu, Zhai, and Gao]{DBLP:conf/iccv/0017LZG23}
Yuan Tian, Guo Lu, Guangtao Zhai, and Zhiyong Gao.
\newblock Non-semantics suppressed mask learning for unsupervised video semantic compression.
\newblock In \emph{{IEEE/CVF} International Conference on Computer Vision, {ICCV} 2023, Paris, France, October 1-6, 2023}, pages 13564--13576. {IEEE}, 2023{\natexlab{a}}.

\bibitem[Tian et~al.(2023{\natexlab{b}})Tian, Yan, Zhai, Chen, and Gao]{DBLP:journals/tip/TianYZCG23}
Yuan Tian, Yichao Yan, Guangtao Zhai, Li Chen, and Zhiyong Gao.
\newblock {CLSA:} {A} contrastive learning framework with selective aggregation for video rescaling.
\newblock \emph{{IEEE} Trans. Image Process.}, 32:\penalty0 1300--1314, 2023{\natexlab{b}}.

\bibitem[Tian et~al.(2024)Tian, Lu, Yan, Zhai, Chen, and Gao]{DBLP:journals/pami/TianLYZCG24}
Yuan Tian, Guo Lu, Yichao Yan, Guangtao Zhai, Li Chen, and Zhiyong Gao.
\newblock A coding framework and benchmark towards low-bitrate video understanding.
\newblock \emph{{IEEE} Trans. Pattern Anal. Mach. Intell.}, 46\penalty0 (8):\penalty0 5852--5872, 2024.

\bibitem[Tsai et~al.(2022)Tsai, Peng, Lin, Tsai, and Lin]{tsai2022stripformer}
Fu-Jen Tsai, Yan-Tsung Peng, Yen-Yu Lin, Chung-Chi Tsai, and Chia-Wen Lin.
\newblock Stripformer: Strip transformer for fast image deblurring.
\newblock In \emph{European conference on computer vision}, pages 146--162. Springer, 2022.

\bibitem[Wang et~al.(2024{\natexlab{a}})Wang, Pan, Wang, Fu, Liang, Wang, Wu, and Liu]{wang2024correlation}
Cong Wang, Jinshan Pan, Wei Wang, Gang Fu, Siyuan Liang, Mengzhu Wang, Xiao-Ming Wu, and Jun Liu.
\newblock Correlation matching transformation transformers for uhd image restoration.
\newblock In \emph{Proceedings of the AAAI Conference on Artificial Intelligence}, pages 5336--5344, 2024{\natexlab{a}}.

\bibitem[Wang et~al.(2020)Wang, Xie, Zhao, and Meng]{wang2020model}
Hong Wang, Qi Xie, Qian Zhao, and Deyu Meng.
\newblock A model-driven deep neural network for single image rain removal.
\newblock In \emph{Proceedings of the IEEE/CVF conference on computer vision and pattern recognition}, pages 3103--3112, 2020.

\bibitem[Wang et~al.(2024{\natexlab{b}})Wang, Liu, and Zhao]{wang2024public}
Songping Wang, Hanqing Liu, and Haochen Zhao.
\newblock Public-domain locator for boosting attack transferability on videos.
\newblock In \emph{2024 IEEE International Conference on Multimedia and Expo (ICME)}, pages 1--6. IEEE, 2024{\natexlab{b}}.

\bibitem[Wang et~al.(2023)Wang, Zhang, Shen, Luo, Stenger, and Lu]{wang2023ultra}
Tao Wang, Kaihao Zhang, Tianrun Shen, Wenhan Luo, Bjorn Stenger, and Tong Lu.
\newblock Ultra-high-definition low-light image enhancement: A benchmark and transformer-based method.
\newblock In \emph{Proceedings of the AAAI Conference on Artificial Intelligence}, pages 2654--2662, 2023.

\bibitem[Wang et~al.(2024{\natexlab{c}})Wang, Zhang, Shao, Luo, Stenger, Lu, Kim, Liu, and Li]{DBLP:journals/ijcv/WangZSLSLKLL24}
Tao Wang, Kaihao Zhang, Ziqian Shao, Wenhan Luo, Bj{\"{o}}rn Stenger, Tong Lu, Tae{-}Kyun Kim, Wei Liu, and Hongdong Li.
\newblock Gridformer: Residual dense transformer with grid structure for image restoration in adverse weather conditions.
\newblock \emph{Int. J. Comput. Vis.}, 132\penalty0 (10):\penalty0 4541--4563, 2024{\natexlab{c}}.

\bibitem[Wang et~al.(2024{\natexlab{d}})Wang, Wang, Zhang, Fan, Wu, Jiang, and Liu]{DBLP:journals/corr/abs-2408-17135}
Yabiao Wang, Shuo Wang, Jiangning Zhang, Ke Fan, Jiafu Wu, Zhengkai Jiang, and Yong Liu.
\newblock Temporal and interactive modeling for efficient human-human motion generation.
\newblock \emph{CoRR}, abs/2408.17135, 2024{\natexlab{d}}.

\bibitem[Wang et~al.(2004)Wang, Bovik, Sheikh, and Simoncelli]{wang2004image}
Zhou Wang, Alan~C Bovik, Hamid~R Sheikh, and Eero~P Simoncelli.
\newblock Image quality assessment: from error visibility to structural similarity.
\newblock \emph{IEEE transactions on image processing}, 13\penalty0 (4):\penalty0 600--612, 2004.

\bibitem[Wang et~al.(2022)Wang, Cun, Bao, Zhou, Liu, and Li]{wang2022uformer}
Zhendong Wang, Xiaodong Cun, Jianmin Bao, Wengang Zhou, Jianzhuang Liu, and Houqiang Li.
\newblock Uformer: A general u-shaped transformer for image restoration.
\newblock In \emph{Proceedings of the IEEE/CVF conference on computer vision and pattern recognition}, pages 17683--17693, 2022.

\bibitem[Wei et~al.(2023)Wei, Wang, and Yan]{wei2023efficient}
Xingxing Wei, Songping Wang, and Huanqian Yan.
\newblock Efficient robustness assessment via adversarial spatial-temporal focus on videos.
\newblock \emph{IEEE Transactions on Pattern Analysis and Machine Intelligence}, 45\penalty0 (9):\penalty0 10898--10912, 2023.

\bibitem[Xiao et~al.(2022)Xiao, Fu, Liu, Wu, and Zha]{xiao2022image}
Jie Xiao, Xueyang Fu, Aiping Liu, Feng Wu, and Zheng-Jun Zha.
\newblock Image de-raining transformer.
\newblock \emph{IEEE Transactions on Pattern Analysis and Machine Intelligence}, 45\penalty0 (11):\penalty0 12978--12995, 2022.

\bibitem[Xiao et~al.(2024{\natexlab{a}})Xiao, Fu, Zhu, Li, Huang, Zhu, and Zha]{DBLP:conf/cvpr/0002FZL00Z24}
Jie Xiao, Xueyang Fu, Yurui Zhu, Dong Li, Jie Huang, Kai Zhu, and Zheng{-}Jun Zha.
\newblock Homoformer: Homogenized transformer for image shadow removal.
\newblock In \emph{{IEEE/CVF} Conference on Computer Vision and Pattern Recognition, {CVPR} 2024, Seattle, WA, USA, June 16-22, 2024}, pages 25617--25626. {IEEE}, 2024{\natexlab{a}}.

\bibitem[Xiao et~al.(2024{\natexlab{b}})Xiao, Lu, and Wang]{DBLP:conf/mm/XiaoLW24}
Zeyu Xiao, Zhihe Lu, and Xinchao Wang.
\newblock P-bic: Ultra-high-definition image moir{\'{e}} patterns removal via patch bilateral compensation.
\newblock In \emph{Proceedings of the 32nd {ACM} International Conference on Multimedia, {MM} 2024, Melbourne, VIC, Australia, 28 October 2024 - 1 November 2024}, pages 8365--8373. {ACM}, 2024{\natexlab{b}}.

\bibitem[Xie et~al.(2024)Xie, Tai, Zhang, Zhang, Zhou, and Yang]{DBLP:journals/corr/abs-2404-01717}
Rui Xie, Ying Tai, Kai Zhang, Zhenyu Zhang, Jun Zhou, and Jian Yang.
\newblock Addsr: Accelerating diffusion-based blind super-resolution with adversarial diffusion distillation.
\newblock \emph{CoRR}, abs/2404.01717, 2024.

\bibitem[Xu et~al.(2022)Xu, Wang, Fu, and Jia]{xu2022snr}
Xiaogang Xu, Ruixing Wang, Chi-Wing Fu, and Jiaya Jia.
\newblock Snr-aware low-light image enhancement.
\newblock In \emph{Proceedings of the IEEE/CVF conference on computer vision and pattern recognition}, pages 17714--17724, 2022.

\bibitem[Yang et~al.(2019)Yang, Tan, Feng, Guo, Yan, and Liu]{yang2019joint}
Wenhan Yang, Robby~T Tan, Jiashi Feng, Zongming Guo, Shuicheng Yan, and Jiaying Liu.
\newblock Joint rain detection and removal from a single image with contextualized deep networks.
\newblock \emph{IEEE transactions on pattern analysis and machine intelligence}, 42\penalty0 (6):\penalty0 1377--1393, 2019.

\bibitem[Yang and Wang(2024)]{yang2024kolmogorov}
Xingyi Yang and Xinchao Wang.
\newblock Kolmogorov-arnold transformer.
\newblock \emph{arXiv preprint arXiv:2409.10594}, 2024.

\bibitem[Ye et~al.(2024)Ye, Chen, Chai, Xing, Qin, Lin, and Zhu]{ye2024learning}
Tian Ye, Sixiang Chen, Wenhao Chai, Zhaohu Xing, Jing Qin, Ge Lin, and Lei Zhu.
\newblock Learning diffusion texture priors for image restoration.
\newblock In \emph{Proceedings of the IEEE/CVF Conference on Computer Vision and Pattern Recognition}, pages 2524--2534, 2024.

\bibitem[Yi et~al.(2021)Yi, Li, Dai, Fang, Zhang, and Zeng]{yi2021structure}
Qiaosi Yi, Juncheng Li, Qinyan Dai, Faming Fang, Guixu Zhang, and Tieyong Zeng.
\newblock Structure-preserving deraining with residue channel prior guidance.
\newblock In \emph{Proceedings of the IEEE/CVF international conference on computer vision}, pages 4238--4247, 2021.

\bibitem[Yu et~al.(2023)Yu, Zhu, Zheng, Huang, Zhou, and Zhao]{DBLP:conf/mm/YuZZHZZ23}
Wei Yu, Qi Zhu, Naishan Zheng, Jie Huang, Man Zhou, and Feng Zhao.
\newblock Learning non-uniform-sampling for ultra-high-definition image enhancement.
\newblock In \emph{Proceedings of the 31st {ACM} International Conference on Multimedia, {MM} 2023, Ottawa, ON, Canada, 29 October 2023- 3 November 2023}, pages 1412--1421. {ACM}, 2023.

\bibitem[Yu et~al.(2024)Yu, Huang, Li, Zheng, Zhu, Zhou, and Zhao]{yu2024empowering}
Wei Yu, Jie Huang, Bing Li, Kaiwen Zheng, Qi Zhu, Man Zhou, and Feng Zhao.
\newblock Empowering resampling operation for ultra-high-definition image enhancement with model-aware guidance.
\newblock In \emph{Proceedings of the IEEE/CVF Conference on Computer Vision and Pattern Recognition}, pages 25722--25731, 2024.

\bibitem[Yu et~al.(2022)Yu, Dai, Li, Ma, Shen, Li, and Qi]{yu2022towards}
Xin Yu, Peng Dai, Wenbo Li, Lan Ma, Jiajun Shen, Jia Li, and Xiaojuan Qi.
\newblock Towards efficient and scale-robust ultra-high-definition image demoir{\'e}ing.
\newblock In \emph{European Conference on Computer Vision}, pages 646--662. Springer, 2022.

\bibitem[Zamir et~al.(2022)Zamir, Arora, Khan, Hayat, Khan, and Yang]{zamir2022restormer}
Syed~Waqas Zamir, Aditya Arora, Salman Khan, Munawar Hayat, Fahad~Shahbaz Khan, and Ming-Hsuan Yang.
\newblock Restormer: Efficient transformer for high-resolution image restoration.
\newblock In \emph{Proceedings of the IEEE/CVF conference on computer vision and pattern recognition}, pages 5728--5739, 2022.

\bibitem[Zhang et~al.(2024{\natexlab{a}})Zhang, Zhang, Gu, Dong, Kong, and Yang]{zhangxformer}
Jiale Zhang, Yulun Zhang, Jinjin Gu, Jiahua Dong, Linghe Kong, and Xiaokang Yang.
\newblock Xformer: Hybrid x-shaped transformer for image denoising.
\newblock In \emph{The Twelfth International Conference on Learning Representations}, 2024{\natexlab{a}}.

\bibitem[Zhang et~al.(2017)Zhang, Zuo, Gu, and Zhang]{zhang2017learning}
Kai Zhang, Wangmeng Zuo, Shuhang Gu, and Lei Zhang.
\newblock Learning deep cnn denoiser prior for image restoration.
\newblock In \emph{Proceedings of the IEEE conference on computer vision and pattern recognition}, pages 3929--3938, 2017.

\bibitem[Zhang et~al.(2024{\natexlab{b}})Zhang, Han, Zhong, Yu, Wu, et~al.]{zhang2024vocapter}
Li Zhang, Zean Han, Yan Zhong, Qiaojun Yu, Xingyu Wu, et~al.
\newblock Vocapter: Voting-based pose tracking for category-level articulated object via inter-frame priors.
\newblock In \emph{ACM Multimedia 2024}, 2024{\natexlab{b}}.

\bibitem[Zhang et~al.(2024{\natexlab{c}})Zhang, Xu, Li, Du, and Wang]{zhang2024catmullrom}
Li Zhang, Mingliang Xu, Dong Li, Jianming Du, and Rujing Wang.
\newblock Catmullrom splines-based regression for image forgery localization.
\newblock In \emph{Proceedings of the AAAI Conference on Artificial Intelligence}, pages 7196--7204, 2024{\natexlab{c}}.

\bibitem[Zhang et~al.(2024{\natexlab{d}})Zhang, Zhong, Wang, Min, Liu, et~al.]{zhangrethinking}
Li Zhang, Yan Zhong, Jianan Wang, Zhe Min, Liu Liu, et~al.
\newblock Rethinking 3d convolution in $\ell_p$-norm space.
\newblock In \emph{The Thirty-eighth Annual Conference on Neural Information Processing Systems}, 2024{\natexlab{d}}.

\bibitem[Zhang et~al.(2025)Zhang, Meng, Zhong, Kong, Xu, Du, Wang, Wang, and Liu]{zhang2025u}
Li Zhang, Weiqing Meng, Yan Zhong, Bin Kong, Mingliang Xu, Jianming Du, Xue Wang, Rujing Wang, and Liu Liu.
\newblock U-cope: Taking a further step to universal 9d category-level object pose estimation.
\newblock In \emph{European Conference on Computer Vision}, pages 254--270. Springer, 2025.

\bibitem[Zhang et~al.(2024{\natexlab{e}})Zhang, Liu, Li, Chen, Liu, Hu, Xiong, Yuan, and Wang]{zhang2024distilling}
Quan Zhang, Xiaoyu Liu, Wei Li, Hanting Chen, Junchao Liu, Jie Hu, Zhiwei Xiong, Chun Yuan, and Yunhe Wang.
\newblock Distilling semantic priors from sam to efficient image restoration models.
\newblock In \emph{Proceedings of the IEEE/CVF Conference on Computer Vision and Pattern Recognition}, pages 25409--25419, 2024{\natexlab{e}}.

\bibitem[Zhang et~al.(2018)Zhang, Isola, Efros, Shechtman, and Wang]{zhang2018unreasonable}
Richard Zhang, Phillip Isola, Alexei~A Efros, Eli Shechtman, and Oliver Wang.
\newblock The unreasonable effectiveness of deep features as a perceptual metric.
\newblock In \emph{Proceedings of the IEEE conference on computer vision and pattern recognition}, pages 586--595, 2018.

\bibitem[Zhang et~al.(2019)Zhang, Li, Li, Zhong, and Fu]{zhang2019residual}
Yulun Zhang, Kunpeng Li, Kai Li, Bineng Zhong, and Yun Fu.
\newblock Residual non-local attention networks for image restoration.
\newblock \emph{arXiv preprint arXiv:1903.10082}, 2019.

\bibitem[Zhao et~al.(2024{\natexlab{a}})Zhao, Cai, Dong, and Hu]{zhao2024wavelet}
Chen Zhao, Weiling Cai, Chenyu Dong, and Chengwei Hu.
\newblock Wavelet-based fourier information interaction with frequency diffusion adjustment for underwater image restoration.
\newblock In \emph{Proceedings of the IEEE/CVF Conference on Computer Vision and Pattern Recognition}, pages 8281--8291, 2024{\natexlab{a}}.

\bibitem[Zhao et~al.(2024{\natexlab{b}})Zhao, Cai, Dong, and Zeng]{zhao2024toward}
Chen Zhao, Weiling Cai, Chenyu Dong, and Ziqi Zeng.
\newblock Toward sufficient spatial-frequency interaction for gradient-aware underwater image enhancement.
\newblock In \emph{ICASSP 2024-2024 IEEE International Conference on Acoustics, Speech and Signal Processing (ICASSP)}, pages 3220--3224. IEEE, 2024{\natexlab{b}}.

\bibitem[Zhao et~al.(2024{\natexlab{c}})Zhao, Cai, Hu, and Yuan]{DBLP:journals/nn/ZhaoCHY24}
Chen Zhao, Weiling Cai, Chengwei Hu, and Zheng Yuan.
\newblock Cycle contrastive adversarial learning with structural consistency for unsupervised high-quality image deraining transformer.
\newblock \emph{Neural Networks}, 178:\penalty0 106428, 2024{\natexlab{c}}.

\bibitem[Zhao et~al.(2024{\natexlab{d}})Zhao, Cai, and Yuan]{zhao2024spectral}
Chen Zhao, Wei-Ling Cai, and Zheng Yuan.
\newblock Spectral normalization and dual contrastive regularization for image-to-image translation.
\newblock \emph{The Visual Computer}, pages 1--12, 2024{\natexlab{d}}.

\bibitem[Zhao et~al.(2024{\natexlab{e}})Zhao, Dong, and Cai]{DBLP:journals/corr/abs-2403-01497}
Chen Zhao, Chenyu Dong, and Weiling Cai.
\newblock Learning {A} physical-aware diffusion model based on transformer for underwater image enhancement.
\newblock \emph{CoRR}, abs/2403.01497, 2024{\natexlab{e}}.

\bibitem[Zhao et~al.(2021)Zhao, Lu, Chen, Yang, and Shamir]{zhao2021deep}
Lin Zhao, Shao-Ping Lu, Tao Chen, Zhenglu Yang, and Ariel Shamir.
\newblock Deep symmetric network for underexposed image enhancement with recurrent attentional learning.
\newblock In \emph{Proceedings of the IEEE/CVF international conference on computer vision}, pages 12075--12084, 2021.

\bibitem[Zheng et~al.(2021{\natexlab{a}})Zheng, Ren, Cao, Hu, Wang, Song, and Jia]{zheng2021ultra}
Zhuoran Zheng, Wenqi Ren, Xiaochun Cao, Xiaobin Hu, Tao Wang, Fenglong Song, and Xiuyi Jia.
\newblock Ultra-high-definition image dehazing via multi-guided bilateral learning.
\newblock In \emph{2021 IEEE/CVF Conference on Computer Vision and Pattern Recognition (CVPR)}, pages 16180--16189. IEEE, 2021{\natexlab{a}}.

\bibitem[Zheng et~al.(2021{\natexlab{b}})Zheng, Ren, Cao, Wang, and Jia]{zhuo2021ultra}
Zhuoran Zheng, Wenqi Ren, Xiaochun Cao, Tao Wang, and Xiuyi Jia.
\newblock Ultra-high-definition image hdr reconstruction via collaborative bilateral learning.
\newblock In \emph{Proceedings of the IEEE/CVF international conference on computer vision}, pages 4449--4458, 2021{\natexlab{b}}.

\bibitem[Zhou et~al.(2023{\natexlab{a}})Zhou, Yang, and Yang]{zhou2023pyramid}
Dewei Zhou, Zongxin Yang, and Yi Yang.
\newblock Pyramid diffusion models for low-light image enhancement.
\newblock \emph{arXiv preprint arXiv:2305.10028}, 2023{\natexlab{a}}.

\bibitem[Zhou et~al.(2024{\natexlab{a}})Zhou, Li, Ma, Zhang, and Yang]{DBLP:conf/cvpr/ZhouLMZY24}
Dewei Zhou, You Li, Fan Ma, Xiaoting Zhang, and Yi Yang.
\newblock {MIGC:} multi-instance generation controller for text-to-image synthesis.
\newblock In \emph{{IEEE/CVF} Conference on Computer Vision and Pattern Recognition, {CVPR} 2024, Seattle, WA, USA, June 16-22, 2024}, pages 6818--6828. {IEEE}, 2024{\natexlab{a}}.

\bibitem[Zhou et~al.(2024{\natexlab{b}})Zhou, Xie, Yang, and Yang]{DBLP:journals/corr/abs-2410-12669}
Dewei Zhou, Ji Xie, Zongxin Yang, and Yi Yang.
\newblock 3dis: Depth-driven decoupled instance synthesis for text-to-image generation.
\newblock \emph{CoRR}, abs/2410.12669, 2024{\natexlab{b}}.

\bibitem[Zhou et~al.(2025)Zhou, Li, Ma, Yang, and Yang]{DBLP:journals/pami/ZhouLMYY25}
Dewei Zhou, You Li, Fan Ma, Zongxin Yang, and Yi Yang.
\newblock {MIGC++:} advanced multi-instance generation controller for image synthesis.
\newblock \emph{{IEEE} Trans. Pattern Anal. Mach. Intell.}, 47\penalty0 (3):\penalty0 1714--1728, 2025.

\bibitem[Zhou et~al.(2023{\natexlab{b}})Zhou, Huang, Guo, and Li]{zhou2023fourmer}
Man Zhou, Jie Huang, Chun-Le Guo, and Chongyi Li.
\newblock Fourmer: An efficient global modeling paradigm for image restoration.
\newblock In \emph{International conference on machine learning}, pages 42589--42601. PMLR, 2023{\natexlab{b}}.

\bibitem[Zhou et~al.(2024{\natexlab{c}})Zhou, Huang, Yan, Hong, Jia, Chanussot, and Li]{zhou2024general}
Man Zhou, Jie Huang, Keyu Yan, Danfeng Hong, Xiuping Jia, Jocelyn Chanussot, and Chongyi Li.
\newblock A general spatial-frequency learning framework for multimodal image fusion.
\newblock \emph{IEEE Transactions on Pattern Analysis and Machine Intelligence}, 2024{\natexlab{c}}.

\bibitem[Zhou et~al.(2024{\natexlab{d}})Zhou, Chen, Pan, Shi, and Yang]{zhou2024adapt}
Shihao Zhou, Duosheng Chen, Jinshan Pan, Jinglei Shi, and Jufeng Yang.
\newblock Adapt or perish: Adaptive sparse transformer with attentive feature refinement for image restoration.
\newblock In \emph{Proceedings of the IEEE/CVF Conference on Computer Vision and Pattern Recognition}, pages 2952--2963, 2024{\natexlab{d}}.

\bibitem[Zou et~al.(2024)Zou, Gao, Yang, and Liu]{zou2024wave}
Wenbin Zou, Hongxia Gao, Weipeng Yang, and Tongtong Liu.
\newblock Wave-mamba: Wavelet state space model for ultra-high-definition low-light image enhancement.
\newblock In \emph{Proceedings of the 32nd ACM International Conference on Multimedia}, pages 1534--1543, 2024.

\end{thebibliography}
}

\end{document}